\documentclass{article}

\PassOptionsToPackage{table,xcdraw}{xcolor}
\PassOptionsToPackage{numbers,square}{natbib}
\usepackage{neurips_2026_arix}
\usepackage[utf8]{inputenc}
\usepackage[T1]{fontenc}
\usepackage{hyperref}
\usepackage{url}
\usepackage{booktabs}
\usepackage{amsfonts}
\usepackage{nicefrac}
\usepackage{microtype}
\usepackage{xcolor}
\usepackage{amsmath}
\usepackage{amssymb}
\usepackage{mathtools}
\usepackage{amsthm}
\usepackage{float}
\usepackage{wrapfig}
\usepackage{subcaption}
\usepackage{multirow}
\usepackage{multicol}
\usepackage{pgfplots}
\usepackage{pgfplotstable}
\usepackage{adjustbox}
\usepackage{graphicx}
\usepackage{caption}
\usepackage{tabularx}
\usepackage{listings}
\usepackage{etoc}
\usepackage{tikz}
\usepackage{algorithm}
\usepackage{algorithmicx}
\usepackage{algpseudocode}
\usepackage{makecell}
\usepackage{siunitx}
\usepackage{threeparttable}
\usepackage[most]{tcolorbox}
\tcbuselibrary{skins,listings,breakable}
\usepackage{inconsolata}
\usepackage[disable,textsize=tiny]{todonotes}

\pgfplotsset{compat=1.18}

\usepackage{amsmath,amsfonts,bm}

\def\eqref#1{equation~\ref{#1}}

\def\1{\bm{1}}

\DeclareMathAlphabet{\mathsfit}{\encodingdefault}{\sfdefault}{m}{sl}
\SetMathAlphabet{\mathsfit}{bold}{\encodingdefault}{\sfdefault}{bx}{n}

\newcommand{\E}{\mathbb{E}}

\newcommand{\Lc}{\mathcal{L}}
\newcommand{\Sc}{\mathcal{S}}
\newcommand{\Dc}{\ensuremath{\mathcal{D}}}

\newtcolorbox{promptbox}[1][]{%
  enhanced,
  breakable,
  sharp corners=south,
  colback=gray!10,
  colframe=green!40!black,
  coltitle=white,
  colbacktitle=green!60!black,
  fonttitle=\bfseries,
  title=Prompt,
  boxrule=0.7pt,
  leftrule=0.7pt,
  rightrule=0.7pt,
  bottomrule=0.7pt,
  toprule=0.7pt,
  titlerule=0pt,
  arc=2mm,
  top=1.0ex,
  bottom=1.0ex,
  left=1.0em,
  right=1.0em,
  attach boxed title to top left={yshift=-2pt, xshift=2pt},
  boxed title style={
    enhanced,
    arc=2mm,
    boxrule=0pt,
    top=0.6ex,
    bottom=0.6ex,
    left=0.9em,
    right=0.9em
  },
  before skip=1ex plus 0.5ex minus 0.2ex,
  after skip=1ex plus 0.5ex minus 0.2ex,
  listing only,
  listing options={
    basicstyle=\ttfamily\small,
    breaklines=true,
    columns=fullflexible,
    showstringspaces=false
  },
  #1
}

\newtheorem{theorem}{Theorem}[section]
\newtheorem{lemma}[theorem]{Lemma}

\theoremstyle{definition}

\theoremstyle{remark}

\title{Sharpness-Guided Group Relative Policy Optimization via Probability Shaping}

\author{%
  Tue Le
  \And
  Linh Ngo Van
  \And
  Duc Ang Nguyen
  \And
  Trung Le
}

\begin{document}

\maketitle

\begin{abstract}
Reinforcement learning with verifiable rewards (RLVR) has become a practical route to improve large language model reasoning, and Group Relative Policy Optimization (GRPO) is a widely used optimizer in this setting. However, RLVR training is typically performed with limited control over generalization. We revisit GRPO through a robustness-based generalization view, where the generalization loss is upper bounded by a combination of the empirical loss and a sharpness surrogate measured by the gradient norm. Building on this perspective, we propose Sharpness-Guided GRPO (GRPO-SG), a simple token-weighted variant of GRPO that downweights tokens likely to cause overly large gradients, reducing sharp updates and stabilizing optimization, thereby improving generalization. Experiments across mathematical reasoning, logic puzzles and tool-augmented question answering show consistent improvements over GRPO, along with smoother gradient-norm trajectories, supporting GRPO-SG as a simple and effective generalization-oriented upgrade to GRPO for RLVR.
\end{abstract}

\section{Introduction}
\label{sec:intro}

Large language models (LLMs) have recently made major strides on challenging reasoning workloads, especially mathematics and programming, with systems such as OpenAI's O-series \citep{jaech2024openai}, DeepSeek-R1 \citep{guo2025deepseek}, Kimi K2 \citep{team2025kimi2}, and Qwen3 \citep{yang2025qwen3} reaching state-of-the-art results. A central driver of these advances is Reinforcement Learning with Verifiable Rewards (RLVR) \citep{gao2024designing, lambert2024tulu, team2025kimi, guo2025deepseek, yang2025qwen3}, which typically performs RL using rule-based outcome signals, for example a binary reward that checks whether the final solution is valid. RLVR offers an effective way to strengthen reasoning by providing stable, task-aligned feedback without relying on expensive human labels or extra reward models. Among the algorithms used in this setting, Group Relative Policy Optimization (GRPO) \citep{shao2024deepseekmath} is widely adopted due to its simplicity and strong empirical performance.


Although recent progress in RLVR has been driven by new algorithms~\citep{yu2025dapo, yue2025vapo, guan2025rstar, zheng2025stabilizing, liu2026gdpogrouprewarddecouplednormalization}, cross-domain applications~\citep{xue2025dancegrpo, liu2025flow, pan2025medvlm}, and unexpected empirical observations ~\citep{wang2025reinforcement, yue2025vapo, zhao2025absolute, chandra2025shape}, most implementations still optimize policy model with limit control over generalization.


In this work, we revisit GRPO through a generalization lens. Theorem~\ref{thm:gen_bound} shows that generalization can be characterized by a robust generalization loss upper bounded by the empirical training loss plus a sharpness term. Using the first-order surrogate in Eq.~(\ref{eq:sharpness_approx}), the sharpness term is measured by the gradient norm. Therefore, smaller gradient norms correspond to less sharp updates, which implies better generalization under this bound. Building on this view, we propose \textbf{GRPO-SG (Sharpness-Guided GRPO)}, a simple modification to GRPO that uses a token-confidence signal to control update magnitudes. Concretely, GRPO-SG assigns each token a weight given by a monotone function of the model's selected-token logit in Eq.~(\ref{eq:token_weight}). These weights are integrated into the standard GRPO surrogate, so the optimization downweights tokens that would otherwise induce sharp, unstable gradients, while preserving learning signal on tokens that the policy is confident about and repeatedly relies on during reasoning. Under the generalization bound above, the objective is direct: use token weights to reduce the gradient-norm term in Eq.~(\ref{eq:sharpness_approx}), thereby improving generalization.

Experiments across diverse settings including logic puzzles, mathematical reasoning, and agentic QA confirm that Sharpness-Guided GRPO consistently outperforms GRPO. In particular, across all these RLVR settings, GRPO-SG delivers substantial gains: on the K\&K logic puzzles benchmark, Qwen2.5-3B improves from \textbf{0.42} to \textbf{0.63} (Table \ref{tab:kk_logic}); in agentic QA, the overall average exact-match accuracy across the benchmark suite nearly doubles from \textbf{13.84} to \textbf{27.29} (Table \ref{tab:agentic}); and on math benchmarks including OlympiadBench and Minerva, we observe average improvements of \textbf{+2.7--6.0\%} (Table \ref{tab:math}). Beyond accuracy, GRPO-SG yields smoother gradient-norm trajectories and more stable training (Figure \ref{fig:grad_norm}), consistent with the sharpness-control mechanism.

In summary, this paper makes three main contributions:
\begin{itemize}
\item We provide a generalization view of GRPO by linking RLVR training to a robust generalization objective and its sharpness surrogate.
\item Guided by this view, we propose \text{GRPO-SG}, a simple GRPO variant that applies confidence-shaped token weights to regulate per-token update magnitude and reduce sharpness.
\item We validate GRPO-SG across diverse RLVR benchmarks, showing consistent gains over GRPO together with smoother gradient-norm trajectories and more stable training.
\end{itemize}

\section{Background}

\subsection{Large Language Models}
\label{subsec:background-llm}
Most \textbf{Large Language Models (LLMs)} adopt a transformer decoder-only architecture~\citep{vaswani2017attention}, parameterized by $\theta \in \mathbb{R}^{d}$ and denoted as $\pi_{\theta}$. The basic unit of an LLM is the token, which may correspond to a word, subword, or character, and is selected from a finite vocabulary $\mathcal{V} = \{v^{1}, \dots, v^{N}\}$ of size $N$. Given a prompt $q$, the model generates a sequence of tokens $o = (o_{1}, \dots, o_{T})$ autoregressively. At each step $t$, the model produces a distribution over $\mathcal{V}$ conditioned on $q$ and the previously generated tokens $o_{<t}$, from which the next token is sampled:
\begin{equation}
    o_{t} \sim \pi_{\theta}(\cdot \mid q, o_{<t}).
\end{equation}
This process continues until an end-of-sequence (EOS) token is produced or the sequence length reaches a predefined maximum $t_{\max}$.

\subsection{Reinforcement Learning with Verifiable Rewards (RLVR)}
\label{subsec:background-rlvr}
\textbf{Reinforcement Learning with Verifiable Rewards} ~\citep{gao2024designing, lambert2024tulu, team2025kimi, guo2025deepseek} provides a framework for optimizing LLMs in domains where correctness can be \emph{deterministically verified}. Typical applications include code generation validated by unit tests, mathematical problem solving with symbolic checkers, or factual QA with exact-match rules or programmatic validators. Unlike preference-based reinforcement learning, RLVR leverages a predefined verifier to produce a reward without requiring human labels. Formally, given a dataset of prompts $\mathcal{D}$, a policy $\pi_{\theta}$, and a reference model $\pi_{\mathrm{ref}}$, the objective is:
\begin{equation}
\label{eq:rlvr-objective}
\begin{aligned}
\max_{\theta}\; 
\mathbb{E}_{q \sim \mathcal{Q},\, o \sim \pi_{\theta}(\cdot \mid q)}
\Big[ R_{\phi}(q,o) \Big] - \beta \,\mathbb{D}_{\mathrm{KL}}\!\left[\pi_{\theta}(o \mid q)\,\Vert\,\pi_{\mathrm{ref}}(o \mid q)\right],
\end{aligned}
\end{equation}
where $R_{\phi}$ is the verifiable reward function and $\beta$ controls the KL regularization. A typical verifiable reward function is defined as:
\begin{equation}
R_{\phi}(q,o) = 
\begin{cases}
1 & \text{if } \mathrm{match}(o, o_{g}), \\[4pt]
-1 & \text{otherwise},
\end{cases}
\end{equation}
where $o_{g}$ is the ground-truth answer and $\mathrm{match}(\cdot,\cdot)\in\{0,1\}$ indicates whether the generated output matches. More generally, $R_{\phi}$ may be graded (e.g., partial credit, length penalty, latency) while remaining \emph{verifiable} by rule-based approaches~\citep{wang2024openhands} or model-based verifiers~\citep{ma2025general}.


\subsection{Group Relative Policy Optimization}
\label{subsec:background-grpo}

\textbf{Group Relative Policy Optimization (GRPO)}, first introduced in \citep{shao2024deepseekmath}, is a policy-gradient algorithm widely used for optimizing the objective in Eq.~(\ref{eq:rlvr-objective}). Unlike the popular PPO algorithm ~\citep{schulman2017proximal}, which requires training an additional value function alongside the LLM, GRPO removes the dependency on an explicit value model. Instead, it estimates advantages by \emph{normalizing rewards within a group} of sampled responses to the same prompt. Specifically, for a prompt $q$ with $G$ sampled responses $\{o_{i}\}_{i=1}^{G}$ and associated scalar rewards $\{r_{i}\}_{i=1}^{G}$, GRPO defines a group-normalized advantage as:
\begin{equation}
\label{eq:grpo_advantages}
\hat{A}_{i,t} = \frac{r_{i} - \mathrm{mean}(\{r_{j}\}_{j=1}^{G})}{\mathrm{std}(\{r_{j}\}_{j=1}^{G})},
\end{equation}
The optimization objective of GRPO can be expressed:
\begin{align}
\mathcal{J}_{\text{GRPO}}(\theta) & =\mathbb{E}_{q\sim\mathcal{Q},\{o_{i}\}_{i=1}^{G}\sim\pi_{\theta_{\mathrm{old}}}(O\mid q)}\Biggl[\frac{1}{\sum_{i=1}^{G}|o_{i}|}\times\nonumber \\
 & \sum_{i=1}^{G}\sum_{t=1}^{|o_{i}|}\Bigl\{\min\!\bigl(r_{i,t}(\theta)\hat{A}_{i,t},\text{clip}(r_{i,t}(\theta),1-\epsilon_{l},1+\epsilon_{h})\hat{A}_{i,t}\bigr)\Biggr]-\beta\mathbb{D}\left(\pi_{\theta}\Vert\pi_{\mathrm{ref}}\right)\Bigr\}\label{eq:grpo_formular}
\end{align}
where $r_{i,t}(\theta) = \tfrac{\pi_{\theta}(o_{i,t} \mid q, o_{i,<t})}{\pi_{\theta_{\mathrm{old}}}(o_{i,t} \mid q, o_{i,<t})}$ is the importance sampling ratio and the KL divergence term. $\pi_{\theta_{\mathrm{old}}}$ denotes the policy used to sample responses, $\pi_{\mathrm{ref}}$ is a frozen reference model, and $\epsilon_l, \epsilon_h$ are the PPO-style clipping threshold hyper-parameters and $\beta$ controls KL regularization. 



\subsection{Token-Level Control in Post-Training}
\label{subsec:token-level-control}

Recent work studies token-level as control signals for LLM alignment and RLVR. In preference optimization, TDPO~\citep{zeng2024tokenleveldirectpreferenceoptimization} reformulates DPO with token-wise preference modeling and forward-KL constraints, TIS-DPO~\citep{liu2024tis} assigns token-level importance weights estimated from the prediction-probability gap of contrastive LLMs, and SWIFT~\citep{letoken} performs self-play fine-tuning with token weights estimated by a stronger teacher model rather than by logits matching.

In RLVR, prior methods use entropy, token probability, or token influence to decide where updates should concentrate. 80/20~\citep{wang20258020rulehighentropyminority} keeps only high-entropy forking tokens, GTPO/GRPO-S~\citep{tan2026gtpogrpostokensequencelevel} reshape rewards with entropy-based signals, AR/Lopti~\citep{yang2025not} rebalance low-probability tokens, Token Hidden Reward~\citep{deng2026tokenhiddenrewardsteering} steers exploration versus exploitation. Table~\ref{tab:rlvr-token-control-comparison} summarizes these methods and highlights how they intervene at different points of the GRPO update.


\begin{table*}[t]
\centering
\scriptsize
\setlength{\tabcolsep}{3pt}
\renewcommand{\arraystretch}{1.18}
\caption{Comparison of token-level control methods for RLVR.}
\label{tab:rlvr-token-control-comparison}
\begin{tabularx}{\textwidth}{@{}>{\raggedright\arraybackslash}p{0.13\textwidth}>{\raggedright\arraybackslash}p{0.23\textwidth}>{\raggedright\arraybackslash}p{0.34\textwidth}>{\raggedright\arraybackslash}X@{}}
\toprule
\textbf{Method} & \textbf{Token signal} & \textbf{Schematic intervention} & \textbf{Key idea} \\
\midrule
80/20 rule~\citep{wang20258020rulehighentropyminority} & Top-$\rho$ high-entropy forking tokens & $m_{i,t}\nabla J_{i,t},\ m_{i,t}=\mathbf{1}[H_{i,t}\ge\tau_\rho^{\mathcal{B}}]$ & Restricts policy-gradient updates to high-entropy forking tokens. \\
\midrule
GTPO~\citep{tan2026gtpogrpostokensequencelevel} & Token entropy $H_{i,t}$ with success/failure partition & $\tilde r_{i,t}^{+}\!\propto r_i+\frac{H_{i,t}}{\sum_{k\in O^+}H_{k,t}};\ \tilde r_{j,t}^{-}\!\propto -1-\frac{1/H_{j,t}}{\sum_{k\in O^-}1/H_{k,t}}$ & Token-level shaped rewards and advantages. \\
\midrule
GRPO-S~\citep{tan2026gtpogrpostokensequencelevel} & Sequence entropy $\bar H_i=\frac{1}{|o_i|}\sum_t H_{i,t}$ & $\hat r_i^\pm=\mathrm{shape}(r_i,\bar H_i)$ \quad $\bar r_i(\theta)=\frac{1}{|o_i|}\sum_t r_{i,t}(\theta)$ & Sequence-level shaped rewards with an averaged sequence ratio. \\
\midrule
AR~\citep{yang2025not} & Token probability $p_{i,t}$ & $\hat A_{i,t}^{\mathrm{AR}}=[\alpha p_{i,t}+(1-\alpha)]\hat A_{i,t}^{\mathrm{old}}$ & Linearly downweights lower-probability tokens in the advantage term. \\
\midrule
Lopti~\citep{yang2025not} & Low-probability set $\mathcal{T}_{\mathrm{low}}=\{t:p_{i,t}\le\eta\}$ & $\hat A^{(1)}=\hat A\,\mathbf{1}[p_{i,t}\le\eta]$ then $\hat A^{(2)}=\hat A\,\mathbf{1}[p_{i,t}>\eta]$ & Sequentially updates low-probability tokens before high-probability tokens. \\
\midrule
Token Hidden Reward~\citep{deng2026tokenhiddenrewardsteering} & Cross-token influence $\mathrm{THR}_{i,t}$ on correct-response likelihood & $\hat A^{\mathrm{THR}(p)}_{i,t}=\mathbf{1}[|\mathrm{THR}_{i,t}|>\tau](1+p\,\operatorname{sign}(\mathrm{THR}_{i,t}))\hat A_{i,t}$ & Uses THR magnitude and sign to steer exploitation ($p>0$) or exploration ($p<0$). \\
\midrule
GRPO-SG (ours) & Selected-token logit $h_{i,t}$ & $\bar r_{i,t}(\theta)=w_{i,t}r_{i,t}(\theta),\ w_{i,t}=\omega(h_{i,t})$ & Sharpness-guided ratio shaping. \\
\bottomrule
\end{tabularx}
\end{table*}

These methods share the view: tokens shouldn't contribute uniformly post-training, differing in signals and targets. GRPO-SG follows, using selected-token logits as confidence signals to shape GRPO via sharpness-guided objectives.


\section{Methodology}
\label{sec:Method}

\subsection{A generalization view of GRPO}\label{sec:gen_view}
We investigate the generalization ability of RL reasoning approaches, such as GRPO. Let  $\mathcal{Q}$ denote the distribution to generate questions $q$ (i.e., $q \sim \mathcal{Q}$). Given $q \sim \mathcal{Q}$, we further denote $\pi_*(\cdot \mid q)$ as the \textit{oracle distribution} to generate ground-truth perfect solutions $o \sim \pi_*(\cdot \mid q)$. Given a parameterized LLM $\pi_\theta$, we want to train it as
\begin{align}\max_{\theta} & \;\mathbb{E}_{q\sim\mathcal{Q},o\sim\pi_{*}\left(\cdot\mid q\right)}\left[\frac{1}{|o|}\sum_{t=1}^{|o|}r^{*}\left(\left[q,o_{<t}\right],o_{t}\right)\pi_{\theta}\left(o_{t}\right)\right] \nonumber -\beta\mathbb{D}_{\text{KL}}\left(\pi_{\theta}\Vert\pi_{ref}\right)\\
\equiv\max_{\theta} & \;\mathbb{E}_{q\sim\mathcal{Q}}\left[\frac{1}{|o|}\sum_{t=1}^{|o|}\mathbb{E}_{o_{\leq t}\sim\pi_{*}^{t}\left(\cdot\mid q\right)}\left[r^{*}\left(\left[q,o_{<t}\right],o_{t}\right)\pi_{\theta}\left(o_{t}\right)\right]\right] -\beta\mathbb{D}_{\text{KL}}\left(\pi_{\theta}\Vert\pi_{ref}\right),\label{eq:gen}
\end{align}
where the regularization term $\mathbb{D}_{\text{KL}}\left(\pi_{\theta}\Vert\pi_{ref}\right)$ preserves $\pi_\theta$ from $\pi_{ref}$, we denote $\pi_{\theta}\left(o_{t}\right)=\pi_{\theta}\left(o_{t}\mid q,o_{<t}\right)$, and $\pi_*^t(\cdot \mid q)$ is the marginal of $\pi_*(\cdot \mid q)$ over $o_{\leq t}$. Here we note that $r_{*}\left(\left[q,o_{<t}\right],o_{t}\right)$ is the optimal reward function obtained from the oracle distribution $\pi_*(\cdot \mid q)$. Moreover, $r_{*}\left(\left[q,o_{<t}\right],o_{t}\right)$ means that given $[q, o_{<t}]$, what the reward to generate $o_t$ is if $o_t \sim \pi_*(o_t \mid q, o_{<t})$. Specifically, the connection between the oracle distribution $\pi_*$ and its optimal reward $r_*$ is as follows:
\begin{align}
\pi_{*}= & \text{argmax}_{\pi}\left\{ \mathbb{E}_{q}\left[\sum_{t=1}^{|o|}\mathbb{E}_{o_{t}\sim\pi\left(\cdot\mid q,o_{<t}\right)}\left[r_{*}\left(\left[q,o_{<t}\right],o_{t}\right)\right]\right]\right.\nonumber \\
 & \left.-\lambda\sum_{t=1}^{|o|}D_{f}\left(\pi\left(\cdot\mid q,o_{<t}\right)\Vert\pi_{old}\left(\cdot\mid q,o_{<t}\right)\right)\right\} ,\label{eq:pi_r}
\end{align}
where $f$ is a convex function. We note that Eq. (\ref{eq:pi_r}) indicates that $\pi_*$ is the distribution that maximizes the reward $r_*([q, o_{<t}], o_t)$ when $o_t \sim \pi_*(\cdot \mid q, o_{<t})$. With some manipulations, we reach the optimal solution in the following lemma.
\begin{lemma} \label{lem:optimal_solution}
The optimal solution of (\ref{eq:pi_r}) is
\begin{equation}
r^{*}\left(\left[q,o_{<t}\right],o_{t}\right)=\lambda f'\left(\frac{\pi_{*}\left(o_{t}\right)}{\pi_{old}\left(o_{t}\right)}\right) + const.\label{eq:r_star}
\end{equation}
Moreover, by choosing $f\left(t\right)=\lambda^{-1}\int_{0}^{t}\frac{\omega\left(x\pi_{old}\left(o_{t}\right)\right)}{\pi_{old}\left(o_{t}\right)}dx
$ for some convex function $\omega(\cdot)$\footnote{This ensures that $f$ defined above is a convex function}, we have
\[
f'\left(t\right)=\lambda^{-1}\frac{\omega\left(t\pi_{old}\left(o_{t}\right)\right)}{\pi_{old}\left(o_{t}\right)}
\]
Finally, by choosing $t=\frac{\pi_{*}\left(o_{t}\right)}{\pi_{old}\left(o_{t}\right)}$, we reach
\begin{align*}
r_{*}\left([q,o_{<t}],o_{t}\right) & =\lambda f'\left(\frac{\pi\left(o_{t}\right)}{\pi_{old}\left(o_{t}\right)}\right)+const =\frac{\omega\left(\pi_{*}\left(o_{t}\right)\right)}{\pi_{old}\left(o_{t}\right)}+const.
\end{align*}
\end{lemma}

The optimization problem in (\ref{eq:gen}) now becomes
\begin{align}
\max_{\theta} \; & \mathbb{E}_\mathcal{Q}\left[\frac{1}{|o|}\sum_{t=1}^{|o|}\mathbb{E}_{o_{\leq t}\sim\pi_{*}^{t}\left(\cdot\mid q\right)}\left[\omega\left(\pi_{*}\left(o_{t}\right)\right)\frac{\pi_{\theta}\left(o_{t}\right)}{\pi_{old}\left(o_{t}\right)}\right]\right] -\beta\mathbb{D}_{\text{KL}}\left(\pi_{\theta}\Vert\pi_{ref}\right).\label{eq:gen_new}
\end{align}

In the following theorem, hinted by GRPO, we change the trajectories $o_{\leq t}$ sampled from the oracle distribution $\pi_*^t$
to those sampled from the old policy $\pi_{old}$, which introduces the shift terms.
\begin{theorem}
    Given a checkpoint LLM $\pi_{old}(\cdot \mid q)$, we can upper-bound the objective function in (\ref{eq:gen_new}) 
    \begin{align} & \mathbb{E}_{q\sim\mathcal{Q}}\left[\frac{1}{|o|}\sum_{t=1}^{|o|}\mathbb{E}_{o_{\leq t}\sim\pi_{old}^{t}\left(\cdot\mid q\right)}\left[\omega\left(\pi_{\theta}\left(o_{t}\right)\right)\frac{\pi_{\theta}\left(o_{t}\right)}{\pi_{old}\left(o_{t}\right)}\right]\right]\nonumber\\
 & +\mathbb{E}_{q\sim\mathcal{Q}}\left[\frac{1}{|o|}\sum_{t=1}^{|o|}\left\{ d\left(\pi_{old}^{t},\pi_{*}^{t}\right)+d'\left(\pi_{\theta}^{t},\pi_{*}^{t}\right)\right\} \right] -\beta\mathbb{D}_{\textup{KL}}\left(\pi_{\theta}\Vert\pi_{ref}\right),\label{eq:gen_up}
\end{align}
where $d$,$d'$ are divergences between two distributions. More details can be found in Appendix \ref{sec:shift}. \label{thm:shift}
\end{theorem}
Ignoring the shift terms, we can rewrite the OP in (\ref{eq:gen_up}) as
\begin{align}\max_{\theta} \; & \mathbb{E}_{\mathcal{Q}}\left[\frac{1}{|o|}\sum_{t=1}^{|o|}\mathbb{E}_{o_{\leq t}\sim\pi_{old}^{t}\left(\cdot\mid q\right)}\left[\omega\left(\pi_{\theta}\left(o_{t}\right)\right)\frac{\pi_{\theta}\left(o_{t}\right)}{\pi_{old}\left(o_{t}\right)}\right]\right] \nonumber -\beta\mathbb{D}_{\text{KL}}\left(\pi_{\theta}\Vert\pi_{ref}\right)\nonumber\equiv\\
\max_{\theta} & \left[\mathbb{E}_{\mathcal{Q},o\sim\pi_{old}\left(\cdot\mid q\right)}\left[\frac{1}{|o|}\sum_{t=1}^{|o|}\omega\left(\pi_{\theta}\left(o_{t}\right)\right)\frac{\pi_{\theta}\left(o_{t}\right)}{\pi_{old}\left(o_{t}\right)}\right]\right] -\beta\mathbb{D}_{\text{KL}}\left(\pi_{\theta}\Vert\pi_{ref}\right).\label{eq:old}
\end{align}

We further rewrite (\ref{eq:old}) using multiple $o_{1:G}\stackrel{iid}{\sim}\pi_{old}\left(\cdot\mid q\right)$:
\begin{align}\max_{\theta}\, & \mathbb{E}_{q\sim\mathcal{Q},o_{1:G}\sim\pi_{old}\left(\cdot\mid q\right)}\left[\frac{1}{\sum_{i=1}^{G}|o_{i}|}\sum_{i=1}^{G}\sum_{t=1}^{|o_{i}|}\omega\left(\pi_{\theta}\left(o_{i,t}\right)\right)\right. \left.\frac{\pi_{\theta}\left(o_{i,t}\right)}{\pi_{old}\left(o_{i,t}\right)}\right]-\beta\mathbb{D}_{\text{KL}}\left(\pi_{\theta}\Vert\pi_{ref}\right).\label{eq:old_mul}
\end{align}
We denote $r_{\theta}\left(o_{i,t}\right)=\frac{\pi_{\theta}\left(o_{i,t}\right)}{\pi_{old}\left(o_{i,t}\right)}$. Moreover, inspired by PPO and GRPO, to encourage learning and unlearning from both well-qualified and unqualified responses $o_i$ by using the advantage function $\hat{A}_{i,t}$, we adopt the optimization problem in (\ref{eq:old_mul}) as
\begin{align}\max_{\theta}\,\mathbb{E}_{q\sim\mathcal{Q},o_{1:G}\sim\pi_{old}\left(\cdot\mid q\right)}\left[\frac{1}{\sum_{i=1}^G |o_{i}|}\sum_{i=1}^{G}\sum_{t=1}^{|o_{i}|}\mathcal{J}\left(o_{i.t}\right)\right]
-\beta\mathbb{D}_{\text{KL}}\left(\pi_{\theta}\Vert\pi_{ref}\right),\label{eq:old_mul_adv}
\end{align}
where we have defined
\begin{align*}
\mathcal{J}\left(o_{i.t}\right) & \coloneqq\min\left\{ \omega\left(\pi_{\theta}\left(o_{i,t}\right)\right)r_{\theta}\left(o_{i,t}\right)\hat{A}_{i,t},\right. \left.\text{clip}\left(\omega\left(\pi_{\theta}\left(o_{i,t}\right)\right)r_{\theta}\left(o_{i,t}\right),1-\epsilon_{l},1+\epsilon_{h}\right)\hat{A}_{i,t}\right\} .
\end{align*}

For the OP in (\ref{eq:old_mul_adv}), we need to optimize the generalization loss over the joined distribution $\mathcal{D}$ including samples $(q, o_1,...,o_G)$ with the pdf $p(q, o_{1:G})$, defined as $\mathcal{Q}\left(q\right)\prod_{i=1}^{G}\pi_{old}\left(o_{i}\mid q\right)$. However, in the actual training, we only rely on a finite training set of $\mathcal{S} =[(q_k, o_{k1},...,o_{kG})]_{k=1,...,N}$ to train an LLM model. This introduces a gap between the generalization and empirical losses. Let us denote \textit{generalization loss} over $\mathcal{D}$  and the \textit{empirical loss} over the training set $\mathcal{S}$ as $\mathcal{L}_{\mathcal{D}}(\pi_\theta)$ and $\mathcal{L}_{\mathcal{S}}(\pi_\theta)$ respectively. To handle the generalization loss, we develop the following theorem.
\begin{theorem}[Generalization via local robustness]\label{thm:gen_bound}
Denote $L = \left(1+\epsilon_{h}\right)G^{1/2}$, $d = |\theta|$, and let $\rho > 0$ be the perturbation radius. With probability at least $1-\delta$ over $\mathcal{S}\sim \mathcal{D}^N$, we have
{\footnotesize
\begin{align}
\mathcal{L}_{\mathcal{D}}(\pi_{\theta})
\le
\max_{\|\theta'-\theta\|_2\le\rho}\mathcal{L}_{\mathcal{S}}(\pi_{\theta'})
+ \frac{4L}{\sqrt{N}}
\left[
\sqrt{
d\log\!\left(
1+\frac{\|\theta\|_{2}^{2}}{\rho^{2}} \cdot
\left(1+\sqrt{\frac{\log N}{d}}\right)^{2}
\right)}
+ 2\sqrt{\log\!\frac{N+d}{\delta}}
+ O(1)
\right].
\label{eq:gen_bound}
\end{align}
}
\end{theorem}

Our proof is adopted from SAM \cite{foret2021sharpnessaware}, but our result is more general because it is applicable to any bounded loss function, thanks to the general PAC-Bayes theorem \cite{PAC_Bayes} we apply. Moreover, a first-order expansion around $\theta$ yields the approximation
\begin{equation}
\max_{\|\theta'-\theta\|\le \rho}\mathcal{L}_{\mathcal{S}}(\pi_{\theta'})
\;\approx\;
\mathcal{L}_{\mathcal{S}}(\pi_\theta) \;+\; \rho\,\|\nabla_\theta \mathcal{L}_{\mathcal{S}}(\pi_\theta)\|_2,
\label{eq:sharpness_approx}
\end{equation}
which highlights two complementary targets for better generalization:
(i) reducing the empirical loss, and (ii) reducing the \emph{sharpness} term measured by the gradient norm.
This develops a token-level view of how GRPO shapes $\|\nabla_\theta \mathcal{L}_{\mathcal{S}}(\pi_\theta)\|_2$,
and proposes a confidence-aware reweighting that directly suppresses sharp directions while preserving learning signal on semantically critical tokens.

\subsection{Sharpness-Guided GRPO for better generalization}\label{sec:trgrpo}

Motivated by Eq.~(\ref{eq:gen_bound})-(\ref{eq:sharpness_approx}), we aim to reduce sharpness by explicitly controlling token-level gradient magnitudes.
Following the generalization development, we introduce an increasing \emph{probability-shaping} function $\omega(\cdot)$ and define a token weight
$w_{i,t} := \omega(\pi_{\theta}(o_{i,t})$.
This yields a weighted GRPO surrogate that remains compatible with PPO/GRPO clipping and advantage learning, while steering optimization toward flatter (less sharp) regions.

We rewrite the empirical loss of (\ref{eq:old_mul_adv}) with $r_{i,t}(\theta) \triangleq \frac{\pi_\theta(o_{i,t}\mid q,o_{i,<t})}{\pi_{\theta_{\mathrm{old}}}(o_{i,t}\mid q,o_{i,<t})}$:
\small{
\begin{align*}
\mathcal{L}_{S}(\pi_{\theta}) & =\mathbb{E}_{\left(q,o_{1:G}\right)\sim\mathcal{S}}\biggl[\frac{1}{\sum_{i=1}^{G}|o_{i}|}\sum_{i=1}^{G}\sum_{t=1}^{|o_{i}|}\min\{w_{i,t}\times\\
 & r_{i,t}(\theta)\hat{A}_{i,t},\text{clip}(w_{i,t}r_{i,t}(\theta),\,1-\epsilon_{l},\,1+\epsilon_{h})\hat{A}_{i,t}\}-\beta\,\mathbb{D}_{\mathrm{KL}}\!\bigl[\pi_{\theta}\,\Vert\,\pi_{\mathrm{ref}}\biggr].
\end{align*}}
\normalsize{}


We now investigate the gradient of the empirical loss $\mathcal{L}_\mathcal{S}(\pi_\theta)$.
\begin{lemma}\label{lem:gradient}
The gradient of empirical loss admits the following form. Define
\[
\bar r_{i,t}(\theta)=w_{i,t}\frac{\pi_{\theta}(o_{i,t})}{\pi_{\mathrm{old}}(o_{i,t})}.
\]
We have
\begin{multline*}
\nabla_{\theta}\mathcal{L}_{\mathcal{S}}(\pi_{\theta})
=
\mathbb{E}_{\left(q,\{o_i\}_{i=1}^{G}\right)\sim\mathcal{S}}
\Biggl[
\frac{1}{\sum_{i=1}^{G}|o_i|}
\sum_{i=1}^{G}\sum_{t=1}^{|o_i|}
w_{i,t}\nabla_{\theta}\log\pi_{\theta}(o_{i,t}) \\
\times
\underset{{\color{red}\gamma_{i,t}}}{\underbrace{
\left(
\frac{\pi_{\theta}(o_{i,t})}{\pi_{\mathrm{old}}(o_{i,t})}\hat{A}_{i,t}
\mathbb{I}_{\mathrm{sharp}}\!\left(\bar r_{i,t}(\theta),\hat{A}_{i,t}\right)
+\beta\frac{\pi_{\mathrm{ref}}(o_{i,t})}{\pi_{\theta}(o_{i,t})}-\beta
\right)}}\Biggr].
\end{multline*}
where we define the sharpness-guided clipping gate $\mathbb{I}_{\text{sharp}}\left(\bar r_{i,t}(\theta),\hat{A}_{i,t}\right)$ as
\begin{equation}
\mathbb{I}_{\mathrm{sharp}}\!\left(\bar r_{i,t}(\theta),\hat{A}_{i,t}\right)
=
\mathbf{1}\!\left[
\bar r_{i,t}(\theta)\hat{A}_{i,t}
\leq
\operatorname{clip}\!\left(\bar r_{i,t}(\theta),1-\epsilon_l,1+\epsilon_h\right)\hat{A}_{i,t}
\right].
\end{equation}
    
\end{lemma}


We represent our LLM as the composition of many layers $f=f_{L}\circ f_{L-1}\circ...\circ f_{l}\circ...\circ f_{1}$ where $f$ maps from the input sequence of tokens $a_0$ to the output sequence of tokens $a_L$. Moreover, the output sequence of tokens $a_L$ is used to compute the logit $h_{i,t}$ for predicting the token $o_{i,t}$. Let's denote $a_{l-1}$ and $a_l$ as the input and output of the $l$-th layer, i.e., $a_{l}=f_{l}\left(a_{l-1},\theta_{l}\right)$ where $\theta_l$ represents the model parameter at the $l$-th layer. To quantify the gradient of update, we develop the following theorem.

\begin{theorem} \label{thm:bounds}
Let us denote the classifier head at the output layer, the Jacobian $\frac{\partial f_{l}\left(a_{l-1},\theta_{l}\right)}{\partial a_{l-1}}$, and the gradient matrix $\frac{\partial f_{l}\left(a_{l-1},\theta_{l}\right)}{\partial\theta_{l}}$ as $W, J_l$, and $G_l$ respectively. Moreover, we further assume that $\sigma_{\textbf{min}}(W) \geq (\alpha^W)^2$, $\sigma_{\textbf{max}}(W) \leq (\beta^W)^2$, $\sigma_{\textbf{min}}(J_l) \geq (\alpha^J_l)^2$, $\sigma_{\textbf{max}}(J_l) \leq (\beta^J_l)^2$, and $\sigma_{\textbf{min}}(G_l) \geq (\alpha^G_l)^2$, $\sigma_{\textbf{max}}(G_l) \leq (\beta^G_l)^2$ where all lower/upper bounds are positive and $\sigma_{\text{min}}(\cdot)$, $\sigma_{\text{max}}(\cdot)$ return the smallest and the largest singular values of a matrix. We can bound the gradient norm $\|g_{i,t}\|_2$ where $g_{i,t}=\gamma_{i,t}w_{it}\nabla_{\theta}\log\pi_{\theta}(o_{i,t})$:
{\small
\begin{equation}
\begin{aligned}
\frac{w_{i,t}(1-\pi_{\theta}(o_{i,t}))}{\sqrt{L}}L_{i,t}
\leq \Vert g_{i,t}\Vert_{2}
\leq
\sqrt{2}\,w_{i,t}(1-\pi_{\theta}(o_{i,t}))U_{i,t},
\end{aligned}
\label{eq:our_grad_norm}
\end{equation}}
where we have defined
\[
L_{i,t} = |\gamma_{i,t}|
\sum_{l=1}^{L}\left(a^{W}a_{l}^{G}\prod_{j=l}^{L}a_{j}^{J}\right), \quad
U_{i,t} = |\gamma_{i,t}|
\sum_{l=1}^{L}\left(b^{W}b_{l}^{G}\prod_{j=l}^{L}b_{j}^{J}\right).
\]
\end{theorem}

With the support of Theorem \ref{thm:bounds}, we develop the lower and upper bounds for $\|\nabla_{\theta}\mathcal{L}_{\mathcal{S}}(\pi_{\theta})\|_2$. 
\begin{theorem} \label{thm:grad_bound}
Let $d = |\theta|$ be the model size. We have

(i) The upper bound of $\|\nabla_{\theta}\mathcal{L}_{\mathcal{S}}(\pi_{\theta})\|_2$ is
\small{
\begin{equation}
\|\nabla_{\theta}\mathcal{L}_{\mathcal{S}}(\pi_{\theta})\|_{2}\leq\mathbb{E}_{\mathcal{S}}\left[\frac{\sqrt{2}}{\sum_{i}^{G}|o_{i}|}\sum_{i=1}^{G}\sum_{t=1}^{|o_{i}|}w_{i,t}\left(1-\pi\left(o_{i,t}\right)\right)U_{i,t}\right]\label{eq:grad_upper}
\end{equation}}
(ii) For any $\delta \in (0,1)$, with probability  at least $1-\delta$, we have
\small{
\begin{multline}
\|\nabla_{\theta}\mathcal{L}_{\mathcal{S}}(\pi_{\theta})\|_{2}\geq 
\left[\frac{1}{|\mathcal{S}|^{2}}\mathbb{E}_{\mathcal{S}}\left[\left(\sum_{i=1}^{G}\sum_{t=1}^{|o_{i}|}\frac{w_{i,t}\left(1-\pi_{\theta}\left(o_{i,t}\right)\right)L_{i,t}}{(\sum_{i=1}^{G}|o_{i}|)^{3/2}\sqrt{L}}\right)^{2}\right]\right.\\
\,\,\,\,\,\,\,\,\,\,\,\,\,\,\,\,-2\epsilon\mathbb{E}_{\mathcal{S}\times\mathcal{S}}\left[\left(\sum_{i=1}^{G}\sum_{t=1}^{|o_{i}|}\frac{w_{i,t}\left(1-\pi_{\theta}\left(o_{i,t}\right)\right)U_{i,t}}{\sum_{i}|o_{i}|}\right)\right.\times\left.\left.\left(\sum_{i=1}^{G}\sum_{t=1}^{|o_{i}^{'}|}\frac{w_{i,t}^{'}\left(1-\pi_{\theta}\left(o_{i,t}^{'}\right)\right)U_{i,t}^{'}}{\sum_{i'}|o_{i}^{'}|}\right)\right]\right]^{1/2} \label{eq:grad_lower}
\end{multline}}
where we define $\epsilon=\sqrt{\frac{1}{cd}\left(\log\left(CK^{2}\right)+\log\frac{1}{\delta}\right)}$ with two positive constants $c, C$ and $K = \sum_{k=1}^N\sum_{i=1}^G |o_{ki}|$.
\end{theorem}

It is worth noting that, because the model size $d = |\theta|$ is usually very high for LLMs, the second term $2\epsilon\mathbb{E}_{\mathcal{S} \times \mathcal{S}}[...]$ is usually negligible, and the lower bound of $\|\nabla_{\theta}\mathcal{L}_{\mathcal{S}}(\pi_{\theta})\|_{2}$ is dominated by the first term. 

According to the upper bound in (\ref{eq:grad_upper}) and the lower bound in (\ref{eq:grad_lower}), we need to set the weights $w_{i,t}$ to minimize $w_{i,t}\left(1-\pi\left(o_{i,t}\right)\right)$, reduce the gradient norm $\|\nabla_{\theta}\mathcal{L}_{\mathcal{S}}(\pi_{\theta})\|_{2}$, and stabilize the gradients.   

We hence instantiate $w_{i,t}$ using a monotone, confidence-aware mapping with stop-gradient to prevent the model from gaming the weights. Inspired by recent advances in token-level preference optimization~\citep{liu2024tis, letoken}, we use the selected-token logit as a monotone proxy and define:
\begin{equation}
\label{eq:token_weight}
w_{i,t} \;=\; \operatorname{clip}\!\Big(\alpha \cdot \big[\sigma\!\big(\tfrac{\operatorname{sg}[h_{i,t}]}{\tau}\big) - \mu\big],\; L,\; U \Big),
\end{equation}
where $h_{i,t}$ denotes the selected-token logit produced by the current policy for token $o_{i,t}$, $\operatorname{sg}[\cdot]$ is the stop-gradient operator, $\tau$ controls the smoothness of the mapping, and $L,U$ bound the token weights.

For low-confidence tokens, $(1-\pi_\theta(o_{i,t}))$ is typically large while $w_{i,t}$ is small; for confident tokens, $(1-\pi_\theta(o_{i,t}))$ is typically small while $w_{i,t}$ is large.
Thus, GRPO-SG stabilizes the product $w_{i,t}(1-\pi_\theta(o_{i,t}))$ across tokens, suppressing extreme token gradients that would otherwise dominate $\|\nabla_\theta \mathcal{L}_{\mathcal{S}}(\pi_\theta)\|_2$.
By Eq.~(\ref{eq:sharpness_approx}), reducing this gradient norm directly reduces sharpness, which tightens the robust empirical bound in Eq.~(\ref{eq:gen_bound}) and improves generalization.

\begin{figure*}[h]
    \centering
    \includegraphics[width=1.0\textwidth]{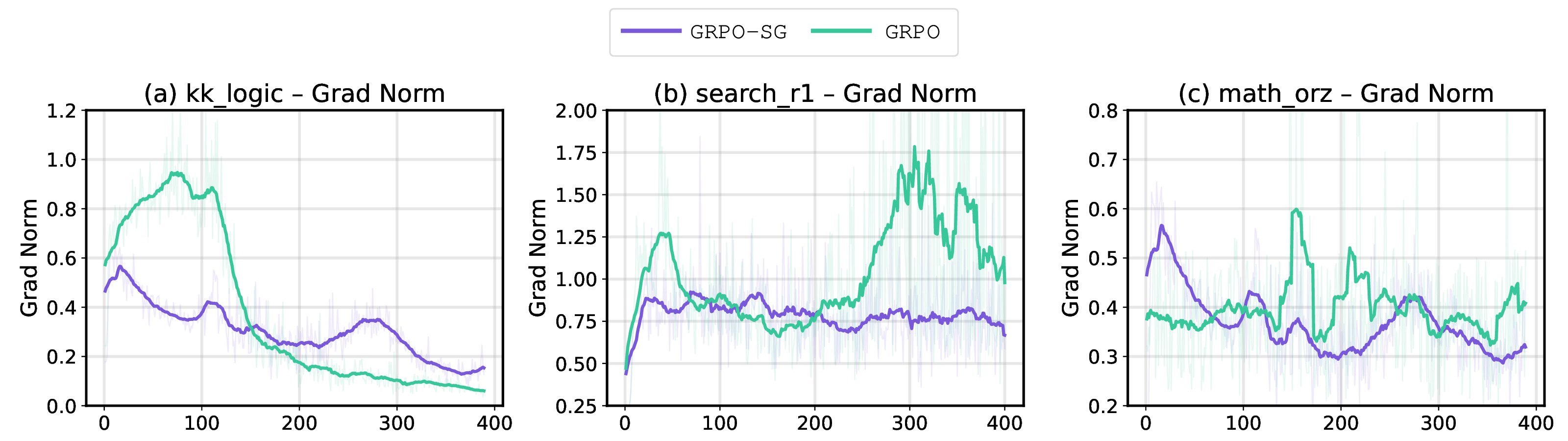}
    \caption{Gradient norm trajectories during training under GRPO vs.\ GRPO-SG across three RLVR settings. GRPO-SG consistently exhibits lower variability and fewer spikes than GRPO, consistent with reduced sharpness in Eq.~(\ref{eq:sharpness_approx}) and the bound in Eq.~(\ref{eq:our_grad_norm}).}
    \label{fig:grad_norm}
\end{figure*}

Eq.~(\ref{eq:sharpness_approx}) suggests that smaller gradient norms correspond to less sharp updates and hence better generalization.
From Theorems~\ref{thm:grad_bound} and \ref{thm:bounds}, GRPO-SG suppresses extreme token gradients through the stabilizing factor $w_{i,t}(1-\pi_\theta(o_{i,t}))$.
To validate this mechanism, we track gradient norms during training under three RLVR settings, as shown in Figure~\ref{fig:grad_norm}. Across all settings, GRPO-SG exhibits smaller fluctuations and fewer outlier spikes than GRPO, consistent with reduced sharpness and improved generalization.

In addition to gradient-norm analysis, we also track training rewards across math reasoning, agentic and logic settings in Appendix Figure~\ref{fig:training_reward}.
GRPO-SG yields higher and more stable reward trajectories than GRPO, indicating more effective learning while simultaneously controlling sharpness.

\section{Experiments}
\label{sec:exp}
We conduct extensive experiments across multiple RLVR benchmarks to assess the effectiveness of GRPO-SG. The results demonstrate that our method consistently outperforms GRPO, delivering stronger reasoning ability and more stable training dynamics.

\begin{table*}[h]
\centering
\caption{Experimental results on math-related datasets (DSR for DeepScaleR and ORZ for Open Reasoner-Zero). The best results are indicated in \textbf{bold}.}
\footnotesize
\setlength{\tabcolsep}{3.5pt}
\renewcommand{\arraystretch}{1.15}
\begin{tabular}{ll*{7}{c}}
\toprule
\textbf{Dataset} & \textbf{Algorithms} &
\makecell[c]{\textbf{Olympiad}\\\textbf{Bench}} &
\makecell[c]{\textbf{Minerva}} &
\makecell[c]{\textbf{MATH}\\\textbf{500}} &
\makecell[c]{\textbf{AMC}\\\textbf{avg@16}} &
\makecell[c]{\textbf{AIME}\\\textbf{avg@16}} &
\makecell[c]{\textbf{Avg.}} &
\makecell[c]{\textbf{Gain}} \\
\midrule
\multicolumn{2}{c}{\textbf{Qwen2.5-7B}} &
28.61 & 17.28 & 64.40 & 23.42 & 4.58 & 27.66 & \\
\midrule
\multirow{5}{*}{\makecell[c]{DSR}}
  & \quad + GRPO & 36.63 & 28.68 & 75.60 & 46.46 & 15.63 & 40.60 & \\
  & \quad + 80/20~\citep{wang20258020rulehighentropyminority} & 37.11 & 27.92 & 75.31 & 44.18 & 14.94 & 39.89 & {\scriptsize \textcolor{red}{$\downarrow$ 1.7\%}} \\
  & \quad + AR~\citep{yang2025not} & 36.94 & 29.11 & 75.92 & 45.23 & 14.72 & 40.38 & {\scriptsize \textcolor{red}{$\downarrow$ 0.5\%}} \\
  & \quad + Lopti~\citep{yang2025not} & 36.52 & 29.56 & 76.23 & 46.26 & 15.45 & 40.80 & {\scriptsize \textcolor{blue}{$\uparrow$ 0.5\%}} \\
  & \quad + GRPO-SG (Ours)  & \textbf{38.48} & \textbf{32.35} & \textbf{79.40} & \textbf{46.84} & \textbf{18.13} & \textbf{43.04} & {\scriptsize \textcolor{blue}{$\uparrow$ 6.0\%}} \\
\midrule
\multirow{5}{*}{\makecell[c]{ORZ}}
  & \quad + GRPO & 38.45 & 27.94 & 76.20 & \textbf{48.72} & 14.79 & 41.22 & \\
  & \quad + 80/20~\citep{wang20258020rulehighentropyminority} & 38.20 & 29.31 & 77.25 & 46.71 & 16.40 & 41.57 & {\scriptsize \textcolor{blue}{$\uparrow$ 0.9\%}} \\
  & \quad + AR~\citep{yang2025not} & 37.95 & 30.22 & 77.71 & 47.19 & 15.52 & 41.72 & {\scriptsize \textcolor{blue}{$\uparrow$ 1.2\%}} \\
  & \quad + Lopti~\citep{yang2025not} & 37.46 & 29.81 & 76.14 & 47.60 & 15.27 & 41.26 & {\scriptsize \textcolor{blue}{$\uparrow$ 0.1\%}} \\
  & \quad + GRPO-SG (Ours)  & \textbf{40.12} & \textbf{30.51} & \textbf{78.60} & 45.48 & \textbf{16.88} & \textbf{42.32} & {\scriptsize \textcolor{blue}{$\uparrow$ 2.7\%}} \\
\bottomrule
\end{tabular}
\label{tab:math}
\end{table*}



\subsection{Experimental Setup}
\label{sec:exp-setup}
To validate the effectiveness and generality of our proposed method, we conduct experiments in three widely used RLVR settings that stress different aspects of reasoning: (i) Math reasoning, (ii) Agentic and (iii) Logic. Additional implementation details, including datasets, evaluation prompts, rewards, training configurations, and evaluation protocols, are provided in Appendix~\ref{app:exp-setup}.

\subsection{Main Results}

We now present the main empirical results. Across math, agentic, and logic RLVR settings, GRPO-SG consistently outperforms GRPO. It also improves over three different other baselines in RLVR: 80/20~\citep{wang20258020rulehighentropyminority}, AR~\citep{yang2025not}, and Lopti~\citep{yang2025not}.

\paragraph{Math-related datasets.}
Across math benchmarks, GRPO-SG gives the most consistent gains in Table~\ref{tab:math}. It improves over GRPO by 2.7--6.0\% on average and achieves the best average performance under both DSR and ORZ. The gains are also broad rather than concentrated in a single benchmark, with improvements on Olympiad Bench, Minerva, and MATH500 under both training recipes. Compared with 80/20, GRPO-SG further improves the average score by +3.15 points on DSR and +0.75 points on ORZ.

\begin{table}[h]
\centering
\caption{Experimental results on the \texttt{K\&K} Logic Puzzles benchmark. The best results are indicated in \textbf{bold}.}
\footnotesize
\setlength{\tabcolsep}{4.0pt}
\renewcommand{\arraystretch}{1.05}
\begin{tabular}{@{}lccccccc@{}}
\toprule
 & \multicolumn{5}{c}{\textbf{Difficulty by Number of People}} &  &  \\
\cmidrule(lr){2-6}
 \textbf{Model} & \textbf{3} & \textbf{4} & \textbf{5} & \textbf{6} & \textbf{7} & \textbf{Avg.} & \textbf{Gain} \\
\midrule
 Qwen2.5-3B-Instruct            & 0.10 & 0.10 & 0.07 & 0.06 & 0.02 & 0.07 &  \\
 \quad + GRPO                   & 0.63 & 0.47 & 0.32 & 0.37 & 0.29 & 0.42 &  \\
 \quad + 80/20~\citep{wang20258020rulehighentropyminority}           & 0.67 & 0.58 & 0.53 & 0.47 & 0.27 & 0.50 & {\scriptsize \textcolor{blue}{$\uparrow$ 19.0\%}} \\
\quad + AR~\citep{yang2025not}           & 0.65 & 0.60 & 0.54 & 0.44 & 0.36 & 0.52 & {\scriptsize \textcolor{blue}{$\uparrow$ 23.8\%}} \\
\quad + Lopti~\citep{yang2025not}         & 0.70 & 0.69 & 0.57 & 0.46 & 0.34 & 0.55 & {\scriptsize \textcolor{blue}{$\uparrow$ 31.0\%}} \\
\quad + GRPO-SG (Ours)                    & \textbf{0.76} & \textbf{0.76} & \textbf{0.61} & \textbf{0.59} & \textbf{0.44} & \textbf{0.63} & {\scriptsize \textcolor{blue}{$\uparrow$ 50.0\%}}
 \\
\midrule
Qwen2.5-7B-Instruct-1M         & 0.24 & 0.18 & 0.11 & 0.10 & 0.04 & 0.13 &  \\
\quad + GRPO                   & 0.92 & 0.90 & 0.74 & 0.59 & 0.60 & 0.75 &  \\
\quad + 80/20~\citep{wang20258020rulehighentropyminority}                   & 0.94 & 0.93 & 0.84 & 0.78 & 0.74 & 0.85 & {\scriptsize \textcolor{blue}{$\uparrow$ 13.3\%}} \\
\quad + AR~\citep{yang2025not}            & 0.95 & \textbf{0.97} & 0.89 & 0.84 & 0.80 & 0.89 & {\scriptsize \textcolor{blue}{$\uparrow$ 18.7\%}} \\
\quad + Lopti~\citep{yang2025not}         & \textbf{0.96} & 0.92 & 0.86 & 0.81 & 0.82 & 0.87 & {\scriptsize \textcolor{blue}{$\uparrow$ 16.0\%}} \\
\quad + GRPO-SG (Ours)                    & 0.95 & 0.95 & \textbf{0.92} & \textbf{0.87} & \textbf{0.84} & \textbf{0.91} & {\scriptsize \textcolor{blue}{$\uparrow$ 21.3\%}}
 \\
\bottomrule

\end{tabular}
\label{tab:kk_logic}
\end{table}

\begin{table*}[h]
\centering
\caption{Experimental results for the agentic task using \texttt{VT-Search} on knowledge-QA benchmarks. $\dagger$ represents in-domain datasets and $\star$ represents out-domain datasets. The best results are indicated in \textbf{bold}.}
\footnotesize
\setlength{\tabcolsep}{3.0pt}
\renewcommand{\arraystretch}{1.15}
\begin{tabular}{lccccccccc}
\toprule
\multirow{2}{*}{\textbf{Model}} 
 & \multicolumn{3}{c}{\textbf{General QA}} 
 & \multicolumn{4}{c}{\textbf{Multi-hop QA}} 
 &  &  \\
\cmidrule(lr){2-4}\cmidrule(lr){5-8}\cmidrule(lr){9-9}
 & \textbf{NQ} & \textbf{TriviaQA} & \textbf{PopQA} 
 & \textbf{HQA} & \textbf{2Wiki} & \textbf{Musique} & \textbf{Bamboogle} 
 & \textbf{Avg.} & \textbf{Gain} \\
\midrule
 Qwen2.5-3B     
 & 0.50 & 1.20 & 0.80 & 0.50 & 1.50 & 0.00 & 0.00 & 0.64 &  \\
\quad + GRPO 
 & 13.50 & 29.40 & 11.50 & 14.50 & 16.30 & 1.30 & 10.40 & 13.84 &  \\
\quad + 80/20~\citep{wang20258020rulehighentropyminority} 
 & 25.95 & 38.55 & 17.28 & 18.54 & 14.30 & 3.61 & \textbf{12.00} & 18.60 & {\scriptsize \textcolor{blue}{$\uparrow$ 34.4\%}} \\
\quad + AR~\citep{yang2025not} 
 & 34.17 & 44.51 & 30.22 & 20.15 & \textbf{22.18} & 4.54 & 7.20 & 23.28 & {\scriptsize \textcolor{blue}{$\uparrow$ 68.2\%}} \\
\quad + Lopti~\citep{yang2025not} 
 & 40.22 & 40.19 & 25.60 & 16.53 & 18.22 & 4.18 & 6.40 & 21.62 & {\scriptsize \textcolor{blue}{$\uparrow$ 56.2\%}} \\
\quad + GRPO-SG (Ours)  
 & \textbf{41.80} & \textbf{52.30} & \textbf{39.10} & \textbf{24.50} & 20.30 & \textbf{5.00} & 8.00 & \textbf{27.29} & {\scriptsize \textcolor{blue}{$\uparrow$ 97.2\%}} \\
\midrule
Qwen3-4B     
 & 31.68 & 60.00 & 38.60 & 27.59 & 14.89 & 6.70 & 23.20 & 28.95 &  \\
\quad + GRPO 
 & 46.48 & 59.84 & 40.39 & 35.54 & 28.87 & 10.26 & 27.20 & 35.21 &  \\
\quad + 80/20~\citep{wang20258020rulehighentropyminority} 
 & 47.94 & 62.67 & 40.27 & 37.52 & 25.66 & 6.51 & 27.20 & 35.40 & {\scriptsize \textcolor{blue}{$\uparrow$ 0.5\%}} \\
\quad + AR~\citep{yang2025not} 
 & \textbf{49.55} & 61.45 & 36.98 & \textbf{38.19} & 23.95 & 8.25 & 34.40 & 36.11 & {\scriptsize \textcolor{blue}{$\uparrow$ 2.6\%}} \\
\quad + Lopti~\citep{yang2025not}
 & 46.21 & 62.09 & 39.11 & 36.74 & 25.94 & \textbf{12.24} & 20.00 & 34.62 & {\scriptsize \textcolor{red}{$\downarrow$ 1.7\%}} \\
\quad + GRPO-SG (Ours) 
 & 48.23 & \textbf{64.11} & \textbf{46.00} & 36.60 & \textbf{30.89} & 10.26 & \textbf{41.60} & \textbf{40.18} & {\scriptsize \textcolor{blue}{$\uparrow$ 14.1\%}} \\
\bottomrule
\end{tabular}
\label{tab:agentic}
\end{table*}

\paragraph{Agentic search.}
The agentic setting in Table~\ref{tab:agentic} is more mixed, since different methods win different retrieval-heavy QA columns. Even so, GRPO-SG delivers the best average on both backbones, moving Qwen2.5-3B from 13.8 to 27.3 and Qwen3-4B from 35.2 to 40.2. On the stronger Qwen3-4B model, it remains competitive on general QA while giving the clearest gains on harder multi-hop columns such as 2Wiki and Bamboogle.

\paragraph{K\&K logic puzzles.} For K\&K logic puzzles, the main pattern is stronger performance as difficulty increases, with GRPO-SG staying ahead in average accuracy in Table~\ref{tab:kk_logic}. It raises the average from 0.42 to 0.63 on Qwen2.5-3B and from 0.75 to 0.91 on Qwen2.5-7B. Lopti is the strongest baseline on 3B and AR on 7B, but neither matches the final average reached by GRPO-SG.


\section{Conclusion}
\label{sec:conclusion}
In this work, we revisit GRPO for RLVR from a generalization perspective and introduce GRPO-SG, a sharpness-guided variant of GRPO that uses token confidence to shape updates and reduce overly sharp gradient steps during RLVR training. Our analysis connects update sharpness to confidence-dependent terms in gradient norms, motivating a simple modification to the GRPO objective. Across diverse settings, GRPO-SG delivers consistent improvements over GRPO. Overall, the results suggest that explicitly controlling update sharpness is an effective way to improve the stability and performance of RL-trained models.

\newpage
\bibliographystyle{unsrtnat}
\bibliography{iclr2026_conference}

\newpage
\appendix
\section{Related Work}
\paragraph{Large-Scale Reasoning Models.} Large language models (LLMs) \citep{lambert2024tulu, gao2024designing, team2025kimi, guo2025deepseek, yang2025qwen3} have recently made substantial advances across a wide range of NLP tasks. A growing line of work now targets stronger performance in reasoning-intensive settings, including mathematics \citep{cobbe2021training,hendrycks2024measuring}, coding \citep{jain2024livecodebench}, and scientific reasoning \citep{rein2024gpqa}. rStar-Math \citep{guan2025rstar} introduces a self-evolving deep-thinking strategy that markedly improves the reasoning ability of smaller LLMs. OpenAI's O-series \citep{jaech2024openai} scales reinforcement learning to train models that solve challenging reasoning problems and achieves state-of-the-art results on several benchmarks.


\paragraph{Reinforcement Learning for Large Language Model.} Before reasoning-centric systems such as OpenAI's O-series \citep{jaech2024openai}, reinforcement learning (RL) was most commonly used through reinforcement learning from human feedback (RLHF) to improve instruction following and preference alignment in large language models (LLMs) \citep{ouyang2022training}. RLHF methods are often grouped into online and offline optimization. Online approaches, such as PPO \citep{schulman2017proximal}, GRPO \citep{shao2024deepseekmath}, and REINFORCE \citep{williams1992simple}, update the policy by sampling model outputs during training and optimizing against immediate reward signals. Offline approaches, including DPO \citep{rafailov2023direct}, SimPO \citep{meng2024simpo}, and KTO \citep{ethayarajh2024kto}, learn from pre-collected preference data provided by annotators or LLMs. Although offline training is typically more efficient, it often underperforms online RL in final capability \citep{tang2024understanding}. More recently, reinforcement learning with verifiable rewards (RLVR) has emerged as a promising route to strengthen LLM reasoning, especially for mathematics and programming. OpenAI o1 \citep{jaech2024openai} provided an early demonstration that RL can scale reasoning ability, and later systems such as DeepSeek-R1 \citep{guo2025deepseek}, Kimi-2 \citep{team2025kimi}, and Qwen3 \citep{yang2025qwen3} have matched or exceeded its performance. In particular, DeepSeek-R1 emphasizes that strong reasoning can arise from outcome-based online RL, notably with GRPO \citep{shao2024deepseekmath}. These results also motivated ``zero RL'' directions that aim to elicit reasoning directly from base models without explicit RL fine-tuning, with follow-up methods including DAPO \citep{yu2025dapo}, VAPO \citep{yue2025vapo}, SimpleRLZoo \citep{zeng2025simplerl}, and Open-Reasoner-Zero \citep{hu2025open}. In parallel, AR-Lopti \citep{yang2025not} shows that low-probability tokens can over-dominate GRPO-style RL updates, and mitigates this via advantage reweighting and low-probability token isolation.

\paragraph{Sharpness Aware Minimization.} 
The relationship between wider, flatter minima and strong generalization has been widely investigated, with both theoretical analyses and empirical evidence reported across many studies \cite{tan2025stabilizing, DBLP:conf/iclr/JiangNMKB20, DBLP:conf/nips/PetzkaKASB21, DBLP:conf/uai/DziugaiteR17, gSAM2022, kwon2021asam}. A common conclusion is that converging to flatter solutions can reduce generalization error and improve robustness under distribution shifts in a range of settings \cite{DBLP:conf/iclr/JiangNMKB20, DBLP:conf/nips/PetzkaKASB21, huang2025learning}. Prior work has also examined how training choices such as batch size, learning rate, gradient covariance, and dropout influence the flatness of the attained minima \cite{DBLP:conf/iclr/KeskarMNST17,Jastrzebski2017ThreeFI, wei2020implicit, deng2025asymptotic}. Sharpness-Aware Minimization (SAM)~\citep{foret2021sharpnessaware} is a more recent optimization approach that targets improved generalization by explicitly accounting for loss-landscape sharpness during training. In particular, SAM optimizes the worst-case loss within a neighborhood of the current parameters, which encourages updates toward flatter regions while maintaining low training loss and better performance on unseen data. SAM has been applied successfully in diverse contexts, including vision~\citep{chen2021vision}, language modeling~\citep{bahri-etal-2022-sharpness}, federated learning~\citep{qu2022generalized, xing2025flexible}, Bayesian neural networks~\citep{vaFlatBNN2023}, domain generalization~\citep{cha2021swad}, multi-task learning~\citep{phan2022stochastic}, and bilevel meta-learning optimization~\citep{abbas2022sharp}

\section{Implementation Details}
\label{app:exp-detail}

\subsection{Experiments setup}
\label{app:exp-setup}

This appendix provides complete details of datasets, prompts, reward design, rollout/training configurations, and evaluation protocols for all three RLVR settings used in this paper: \textbf{Math reasoning}, \textbf{Agentic (VT-Search)} and \textbf{Logic (K\&K)}.

\subsubsection{Math-related dataset}
For \textbf{mathematical} reasoning tasks, where correctness can be verified automatically against gold-standard answers. Following~\citep{yang2025not}, we train on datasets that combine symbolic manipulation and arithmetic word problems, using a binary reward signal: $1$ if the final boxed answer matches the reference solution and $0$ otherwise. The training corpus includes diverse math reasoning prompts designed to encourage structured derivations rather than direct guessing. For evaluation, we adopt multiple benchmarks that are standard in math-focused RLVR research: \texttt{Olympiad Bench}~\citep{he2024olympiadbench}, \texttt{Minerva}~\citep{lewkowycz2022solving}, \texttt{MATH-500}~\citep{hendrycks2021measuring}, \texttt{AMC 2022-2023} and \texttt{AIME 2024}. For the first three benchmarks, evaluation is conducted using greedy decoding. For AMC and AIME, consistent with standard practice, we generate 16 responses per question and report the mean accuracy across these samples (avg@16).

Inspired by~\citep{yang2025not}, we carry out experiments on two math-focused training datasets, DSR-Uniform ($10,000$ problems, evenly covering difficulty levels) and ORZ ($57,000$ problems). In line with prior studies, we adopt \texttt{Qwen2.5-7B}~\citep{yang2025qwen3} as the base model. In this setting, no instruction-tuned templates are applied; instead, we employ a simple prompt directly.

\begin{promptbox}
\textcolor{blue}{\{problem\}} Let's think step by step and output the final answer within \textbackslash\textbackslash boxed\{\}.
\end{promptbox}

LLMs without post-training generally struggle to follow strict output formats. Consequently, format-related signals are not included during training. Moreover, math tasks usually admit only a single correct solution, making partial credit unnecessary. Hence, a binary reward scheme is sufficient: the model receives a reward of $1$ for a correct answer and $0$ otherwise.

\subsubsection{Agentic: VT-Search (knowledge-augmented QA)}

We next consider \textbf{agentic} reasoning tasks, we examine an agentic setting that requires knowledge-intensive reasoning augmented with retrieval tools. Following the \textbf{Search-R1}~\citep{jin2025search} configuration from the \textbf{VerlTool} framework~\citep{jiang2025verltool}, the model interacts with an external tool server providing a retriever and other utilities such as Python or SQL. Training data consists of open-domain QA datasets where each instance requires grounding reasoning steps with retrieved evidence. The reward is based on exact-match correctness of the final answer, while intermediate tool calls are masked in the policy loss to avoid leakage of supervision. For evaluation, we follow the VerlTool benchmark and report exact-match accuracy on both General Q\&A benchmarks (\texttt{NQ}~\citep{kwiatkowski2019natural}, \texttt{TriviaQA}~\citep{joshi2017triviaqa}, \texttt{PopQA}~\citep{mallen2022not}) and multi-hop Q\&A benchmarks (\texttt{HotpotQA}~\citep{yang2018hotpotqa}, \texttt{2Wiki}~\citep{ho2020constructing}, \texttt{MuSiQue}~\citep{trivedi2021musique}, \texttt{Bamboogle}~\citep{press2022measuring}).

Building on prior work~\citep{jin2025search, song2025r1}, an E5 retriever~\citep{wang2022text} is employed with the 2018 Wikipedia dump~\citep{karpukhin2020dense} as the indexed corpus. The agent alternates between retrieval operations and reasoning steps to form complete answers. We adopt \texttt{Qwen2.5-3B}~\citep{yang2025qwen3} and \texttt{Qwen3-4B-Instruct-2507}~\citep{yang2025qwen3} as the base models.

For this task, we use accuracy as the main reward, defined as:

\begin{equation}
    R_{\text{search}}(\mathbf{x}, \mathbf{y}) = \begin{cases}
1 & \text{if } \text{match}(\mathbf{y}, \mathbf{y}_g) \\
-1 & \text{otherwise}
\end{cases}
\end{equation}

\subsubsection{Logic: Knights \& Knaves (K\&K)}
For the \textbf{logic} data, following~\citep{yang2025not}, we adopt the \texttt{K\&K} (Knights and Knaves) logic puzzles introduced in~\citep{xie2024memorization}. Each puzzle describes a set of people who always lie or always tell the truth, and the task is to determine their identities given a set of statements. For training, we use the same dataset splits as in~\citep{yang2025not}, which include synthetic instances of increasing difficulty from 3 to 7 (measured by the number of people per puzzle). The reward function checks both the output format (requiring explicit \texttt{<think>} and \texttt{<answer>} tags) and the correctness of the final assignment. During evaluation, we follow the official harness and report accuracy across all difficulty levels, as well as breakdowns by puzzle size. This setting emphasizes step-by-step deductive reasoning and tests whether the model can align reasoning traces with verifiable logical consistency.

Following~\citep{xie2025logic, yang2025not}, we initialize from instruction-tuned models, \texttt{Qwen2.5-3B-Instruct}~\citep{yang2025qwen3} and \texttt{Qwen2.5-7B-Instruct-1M}~\citep{yang2025qwen2}. The tailored prompt designed for the LLMs is provided below.

\begin{promptbox}
system\textbackslash n You are a helpful assistant. The assistant first thinks about the reasoning
process in the mind and then provides the user with the answer. The reasoning process
and answer are enclosed within <think></think> and <answer></answer> tags, respectively, i.e., <think> reasoning process here </think><answer> answer here </answer>.
Now the user asks you to solve a logical reasoning problem. After thinking, when you
finally reach a conclusion, clearly state the identity of each character within <answer></answer> tags. i.e., <answer> (1) Zoey is a knight\textbackslash n (2) ... </answer>.\textbackslash n user\textbackslash n\textcolor{blue}{\{problem\}}\textbackslash n assistant\textbackslash n<think>
\end{promptbox}

To promote chain-of-thought (CoT) reasoning in LLMs,~\citep{xie2025logic} introduces a reward function with two main components, as shown in Table~\ref{tab:kk-reward}. The output is judged as fully correct if the model generates CoT reasoning wrapped inside \texttt{<think>}...\texttt{</think>} tags and the final prediction enclosed within \texttt{<answer>}...\texttt{</answer>} tags.

\begin{table}[h]
\centering
\small
\caption{Reward design for Logic (K\&K).}
\label{tab:kk-reward}
\setlength{\tabcolsep}{6.5pt}
\renewcommand{\arraystretch}{1.15}
\begin{tabular}{lcc}
\toprule
 & \textbf{Format Reward} & \textbf{Answer Reward} \\
\midrule
Completely Correct  & 1    &  2 \\
Partially Correct   & $-1$ & $-1.5$ \\
Completely Wrong    & $-1$ & $-2$ \\
\bottomrule
\end{tabular}
\end{table}

\subsection{Hyperparameters}

The main GRPO hyperparameter settings are summarized in Table~\ref{tab:hp-grpo-core}. The \emph{clip-higher} technique from DAPO~\citep{yu2025dapo} is used to stabilize entropy and avoid collapse. For token importance estimation (Eq.~\ref{eq:token_weight}), the configuration uses the selected-token logit together with ($\alpha=2.0$) and ($\mu=0.25$), with clipping bounds ($L=0.9$) and ($U=1.4$), and the scaling parameter fixed to ($\tau=9.0$). All experiments run on 8 NVIDIA H100 GPUs.

\begin{table*}[h]
\centering
\caption{Key hyperparameters for GRPO training.}
\label{tab:hp-grpo-core}
\setlength{\tabcolsep}{5.5pt}
\renewcommand{\arraystretch}{1.12}
\begin{tabular}{lccc}
\toprule
\textbf{Hyperparameter} & \textbf{Math reasoning} & \textbf{Agentic (Search--R1)} & \textbf{Logic (K\&K)} \\
\midrule
Group size per prompt $G$ & 8 & 8 & 8 \\
Sampling temperature & 1.0 & 0.8 & 0.7 \\
Max response length & 4096 & 4096 & 4096 \\
Optimizer / LR & AdamW / $1\times 10^{-6}$ & AdamW / $1\times 10^{-6}$ & AdamW / $1\times 10^{-6}$ \\
KL penalty coefficient & 0.001 & 0.001 & 0.001 \\
PPO clip ratio (low / high) & 0.20 \, / \, 0.24 & 0.20 \, / \, 0.24 & 0.20 \, / \, 0.24 \\
Mini-batch size & 128 & 32 & 64 \\
Micro-batch size (updates) & 512 & 256 & 256 \\
\bottomrule
\end{tabular}
\end{table*}


\section{Additional Experiment Results}

\subsection{Hyperparameter sensitivity analysis}
In Eq.~\ref{eq:token_weight}, the token weight is determined by five hyperparameters: $\alpha$, $\mu$, $L$, $U$, and $\tau$. Among these, $\alpha$ and $\mu$ mainly control the global offset and slope of the weight distribution after passing the selected-token logit through the sigmoid map. In our runs, these settings keep most weights centered near $1.0$ after clipping, while still allowing enough variation to emphasize or de-emphasize tokens depending on model confidence.

We then validate the two remaining hyperparameters, $\tau$ and the clipping range $(L,U)$. Table~\ref{tab:hyperparam_sensitivity} reports results on the \texttt{K\&K} Logic Puzzles benchmark using Qwen2.5-3B-Instruct. For $\tau$, we sweep across ${[0.5,2.0,7.0,9.0,10.0,20.0]}$. As $\tau$ increases, the weighting function becomes flatter and the weights tend to be nearly identical, making our method less effective. In contrast, very small $\tau$ produces overly sharp and highly disparate weights, which can amplify per-token variance and destabilize training. We observe the best performance at a moderate value around $\tau=9.0$.

For $(L,U)$, we consider four ranges: $(1.0,1.2)$, $(0.9,1.4)$, $(0.8,1.5)$, and $(0.5,1.5)$. The choice of clipping bounds has only a mild effect, and all settings achieve comparable performance. This suggests that the precise clipping range mainly acts as a safeguard against extreme values rather than a critical tuning factor.

\begin{table*}[h]
  \centering
  \small
  \caption{Hyperparameter sensitivity of token weight estimation on Qwen2.5-3B-Instruct on the \texttt{K\&K} Logic Puzzles benchmark.}
  \label{tab:hyperparam_sensitivity}
  \setlength{\tabcolsep}{6pt}
  \renewcommand{\arraystretch}{1.1}
  \begin{tabular}{lc|cccccc}
    \toprule
    Parameter & Value & 3 & 4 & 5 & 6 & 7 & Avg. \\
    \midrule
    \multirow{6}{*}{$\tau$} 
      & 0.5 & 0.63 & 0.48 & 0.35 & 0.31 & 0.20 & 0.39 \\
      & 2.0 & 0.64 & 0.49 & 0.37 & 0.33 & 0.19 & 0.40 \\
      & 7.0 & 0.67 & 0.62 & 0.41 & 0.37 & 0.24 & 0.46 \\
      & 9.0 & 0.76 & 0.76 & 0.61 & 0.59 & 0.44 & 0.63 \\
      & 10.0 & 0.77 & 0.79 & 0.58 & 0.57 & 0.37 & 0.62 \\
      & 20.0 & 0.74 & 0.68 & 0.51 & 0.46 & 0.35 & 0.55 \\
    \midrule
    \multirow{4}{*}{($L,U$)} 
      & (1.0, 1.2) & 0.77 & 0.76 & 0.60 & 0.58 & 0.40 & 0.62 \\
      & (0.9, 1.4) & 0.76 & 0.76 & 0.61 & 0.59 & 0.44 & 0.63 \\
      & (0.8, 1.5) & 0.74 & 0.75 & 0.57 & 0.62 & 0.42 & 0.62 \\
      & (0.5, 1.5) & 0.70 & 0.65 & 0.59 & 0.60 & 0.44 & 0.60 \\
    \bottomrule
  \end{tabular}
\end{table*}

\subsection{Computational Costs}
\label{subsec:comp-costs}

Table~\ref{tab:lopti-cost} reports the computational overhead of applying GRPO-SG compared to the GRPO baseline on the \texttt{K\&K} Logic Puzzles dataset. The only additional operation required by GRPO-SG is the computation of token-level weights. Importantly, these weights are derived directly from the model's own selected-token logits and therefore do not require any auxiliary extra forward passes. As a result, peak GPU memory usage remains essentially unchanged between GRPO and GRPO-SG across both the \texttt{Qwen2.5-3B-Instruct} and \texttt{LLaMA3.1-8B-Instruct} backbones.

\begin{table}[h]
\centering
\caption{Computational cost comparison of GRPO and GRPO-SG on K\&K Logic Puzzle Dataset. Training time is reported in minutes per sample; peak GPU memory is reported in GB.}
\label{tab:lopti-cost}
\small
\setlength{\tabcolsep}{2pt}
\renewcommand{\arraystretch}{1.15}
\begin{tabular}{lcccc}
\toprule
\multirow{2}{*}{\textbf{}} & \multicolumn{2}{c}{\textbf{Qwen2.5-3B}} & \multicolumn{2}{c}{\textbf{LLaMA3.1-8B}} \\
\cmidrule(lr){2-3} \cmidrule(lr){4-5}
 & \textbf{GRPO} & \textbf{GRPO-SG} & \textbf{GRPO} & \textbf{GRPO-SG} \\
\midrule
Training Time/Sample & 271.4 & 286.0 & 678.5 & 722.2 \\
Peak Mem & 459.7 & 460.2 & 581.2 & 580.6 \\
\bottomrule
\end{tabular}
\end{table}

In terms of runtime, GRPO-SG does introduce a modest increase in training time per sample (e.g., 271.4 vs.\ 286.0 minutes on \texttt{Qwen2.5-3B-Instruct}, and 678.5 vs.\ 722.2 minutes on \texttt{LLaMA3.1-8B-Instruct}). However, this overhead is relatively minor compared to the substantial performance improvements reported in Section~\ref{sec:exp}. Together, these results confirm that GRPO-SG delivers consistent gains in reasoning performance without imposing significant additional computational costs.

\subsection{Limitations}
\label{subsec:limitations}

A key limitation of GRPO-SG is the extra compute from token sharpness control. Each update estimates and applies weights to all generated tokens, slightly increasing per-step cost over standard GRPO. However, as shown in Appendix~\ref{subsec:comp-costs}, the overhead is acceptable in practice and does not limit scalability to larger models or datasets.

\subsection{Training Reward Trajectories}
\label{subsec:training-reward}

\begin{figure*}[h]
        \centering
        \includegraphics[width=1.0\textwidth]{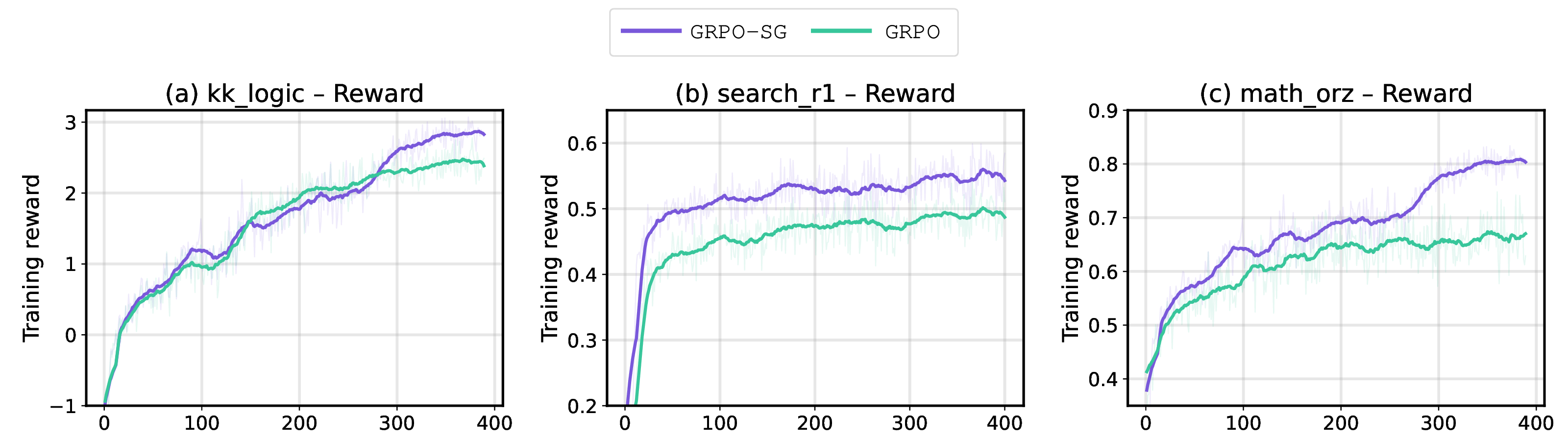}
        \caption{Training reward trajectories during training under GRPO vs.\ GRPO-SG across three RLVR settings. GRPO-SG achieves higher reward while also exhibiting lower sharpness as reflected by gradient norms.}
        \label{fig:training_reward}
\end{figure*}

We also track training rewards across math reasoning, agentic and logic settings. Figure~\ref{fig:training_reward} shows that GRPO-SG consistently achieves higher training rewards throughout optimization.

\subsection{Cross-Model Validation}

Beyond the Qwen family used in previous experiments, which is widely adopted in recent reasoning work, we further validate GRPO-SG on other model families, including \texttt{Mistral} and \texttt{LLaMA}. In addition, we test on the base variant \texttt{Qwen2.5-3B-Base}
rather than the commonly used instruction-tuned variant. All evaluations follow the K\&K logic setting, with results summarized in Table~\ref{tab:var_model}.

\begin{table*}[t]
\centering
\caption{Experimental results on the \texttt{K\&K} Logic Puzzles benchmark on various models. The best results are indicated in \textbf{bold}.}
\setlength{\tabcolsep}{7pt}
\renewcommand{\arraystretch}{1.15}
\begin{tabular}{lccccccc}
\toprule
 & \multicolumn{5}{c}{\textbf{Difficulty by Number of People}} &  &  \\
\cmidrule(lr){2-6}
\textbf{Model} & \textbf{3} & \textbf{4} & \textbf{5} & \textbf{6} & \textbf{7} & \textbf{Avg.} &  \\
\midrule
 Qwen2.5-3B-Instruct & 0.10 & 0.10 & 0.07 & 0.06 & 0.02 & 0.07 &  \\
\quad + GRPO        & 0.63 & 0.47 & 0.32 & 0.37 & 0.29 & 0.42 &  \\
\quad + GRPO-SG     & \textbf{0.76} & \textbf{0.76} & \textbf{0.61} & \textbf{0.59} & \textbf{0.44} & \textbf{0.63} & {\scriptsize \textcolor{blue}{$\uparrow$ 50.0\%}} \\
\midrule
Qwen2.5-7B-Instruct-1M & 0.24 & 0.18 & 0.11 & 0.10 & 0.04 & 0.13 & \\
\quad + GRPO        & 0.92 & 0.90 & 0.74 & 0.59 & 0.60 & 0.75 & \\
\quad + GRPO-SG     & \textbf{0.95} & \textbf{0.95} & \textbf{0.92} & \textbf{0.87} & \textbf{0.84} & \textbf{0.91} & {\scriptsize \textcolor{blue}{$\uparrow$ 21.3\%}} \\
\midrule
 Qwen2.5-3B-Base & 0.14 & 0.04 & 0.02 & 0.01 & 0.02 & 0.05 & \\
\quad + GRPO        & 0.60 & 0.54 & 0.43 & 0.38 & \textbf{0.28} & 0.45 & \\
\quad + GRPO-SG     & \textbf{0.68} & \textbf{0.62} & \textbf{0.44} & \textbf{0.47} & 0.26 & \textbf{0.49} & {\scriptsize \textcolor{blue}{$\uparrow$ 8.9\%}} \\
\midrule
 Mistral-7B-Instruct-v0.3 & 0.05 & 0.01 & 0.00 & 0.02 & 0.00 & 0.02 & \\
\quad + GRPO        & 0.29 & 0.16 & 0.09 & 0.11 & 0.03 & 0.14 & \\
\quad + GRPO-SG     & \textbf{0.47} & \textbf{0.27} & \textbf{0.18} & \textbf{0.15} & \textbf{0.08} & \textbf{0.23} & {\scriptsize \textcolor{blue}{$\uparrow$ 64.3\%}} \\
\midrule
 LLaMA3.1-8B-Instruct & 0.08 & 0.00 & 0.00 & 0.00 & 0.00 & 0.02 & \\
\quad + GRPO        & \textbf{0.92} & 0.92 & 0.83 & 0.79 & 0.80 & 0.85 & \\
\quad + GRPO-SG     & 0.89 & \textbf{0.95} & \textbf{0.88} & \textbf{0.87} & \textbf{0.81} & \textbf{0.88} & {\scriptsize \textcolor{blue}{$\uparrow$ 3.5\%}} \\
\bottomrule
\end{tabular}
\label{tab:var_model}
\end{table*}

Table~\ref{tab:var_model} shows that GRPO-SG consistently improves over GRPO across all tested backbones, indicating that our method generalizes beyond Qwen-Instruct models. This provides further evidence that the proposed token-control strategy is broadly applicable and not limited to a single model family.

\subsection{Word Clouds for High- and Low-Probability Tokens}

\begin{figure}[H]
  \centering
  \begin{subfigure}{0.48\textwidth}
    \includegraphics[width=\linewidth]{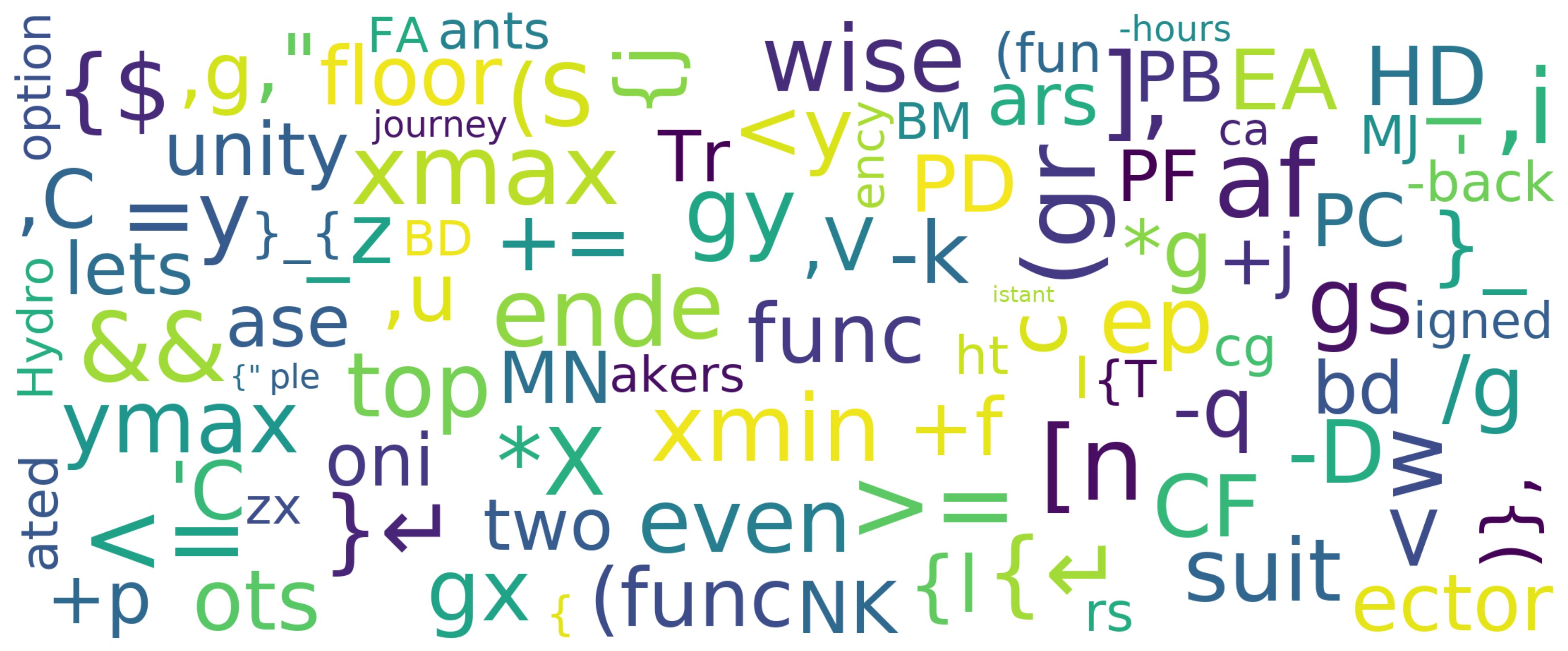}
    \caption{Frequent tokens with the highest average probability.}
    \label{fig:wc-high}
  \end{subfigure}\hfill
  \begin{subfigure}{0.48\textwidth}
    \includegraphics[width=\linewidth]{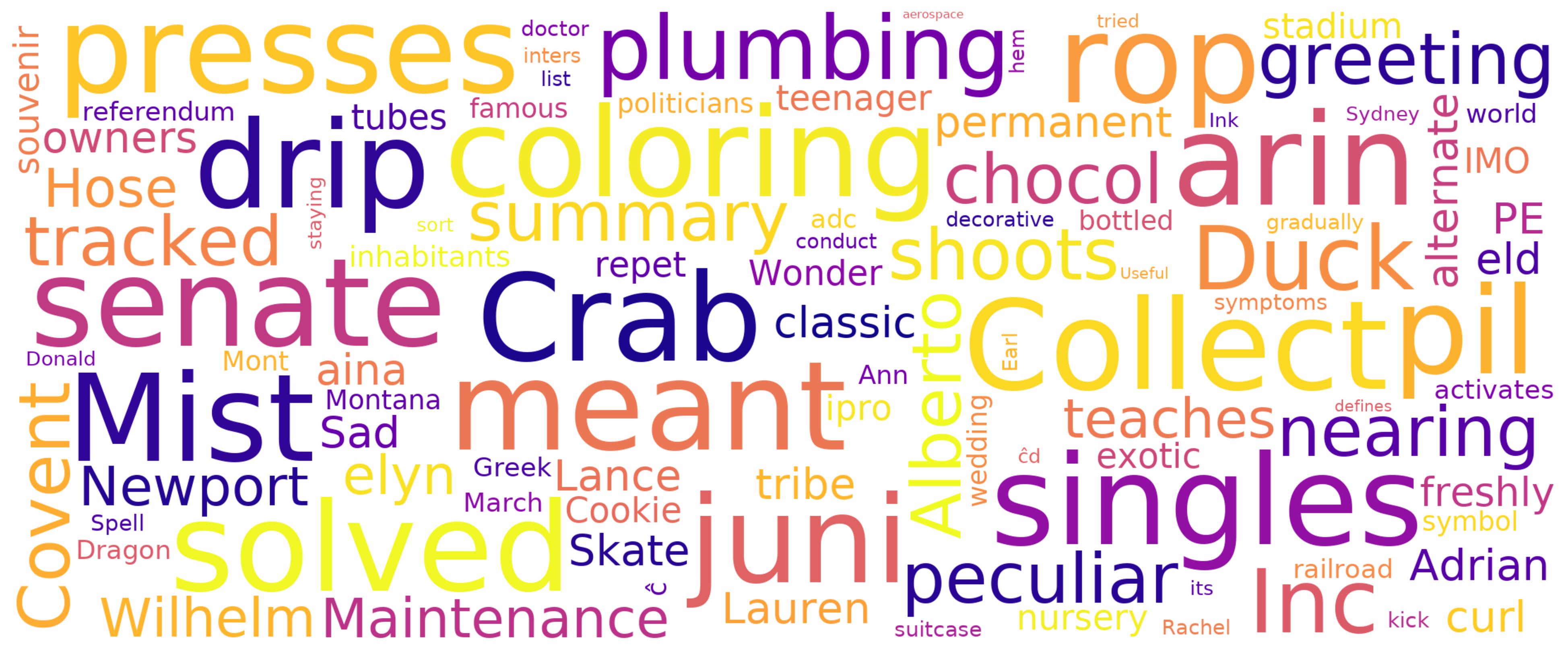}
    \caption{Frequent tokens with the lowest average probability.}
    \label{fig:wc-low}
  \end{subfigure}
  \caption{Word clouds of the top 100 high- vs. low-probability tokens selected from frequently occurring words. High-probability tokens (left) primarily consist of mathematical and logical operators, brackets, and variable names, where even small errors can invalidate an entire solution, whereas low-probability tokens (right) mostly consist of generic content words that are less critical.
}
  \label{fig:wc-both}
\end{figure}

Our method assigns each token a weight that increases with its selected-token logit under the current policy, which acts as a confidence signal for the sampled token. To check whether this weighting mechanism aligns with semantic importance, we visualize the top 100 frequent high-probability tokens and low-probability tokens ranked by their average model probability \(\bar\pi(o_{t})=\tfrac{1}{\#\mathrm{occ}(o_t)}\sum_{o_t} \pi_\theta(o_t)\) (Figures~\ref{fig:wc-high}--\ref{fig:wc-low}), where $\#\mathrm{occ}(o_t)$ denotes the number of occurrences of token $o_t$. Figure~\ref{fig:wc-high} shows that high-probability tokens concentrate on mathematical and logical \emph{structure}---operators, brackets, variable names, and formatting markers---where even a single error can break the entire solution, whereas low-probability tokens are mostly generic content words (e.g., ``output,'' ``particular,'' or ``location'') that contribute less to the core reasoning and are more easily replaced without changing meaning. This creates a clear tension: standard GRPO naturally magnifies low-probability tokens (via larger gradients) even though high-probability tokens carry the most critical signal for correctness. This analysis further supports our method.


\subsection{Ablation on Confidence-Based Token Weighting}

\begin{figure*}[h]
    \centering
    \includegraphics[width=1.0\textwidth]{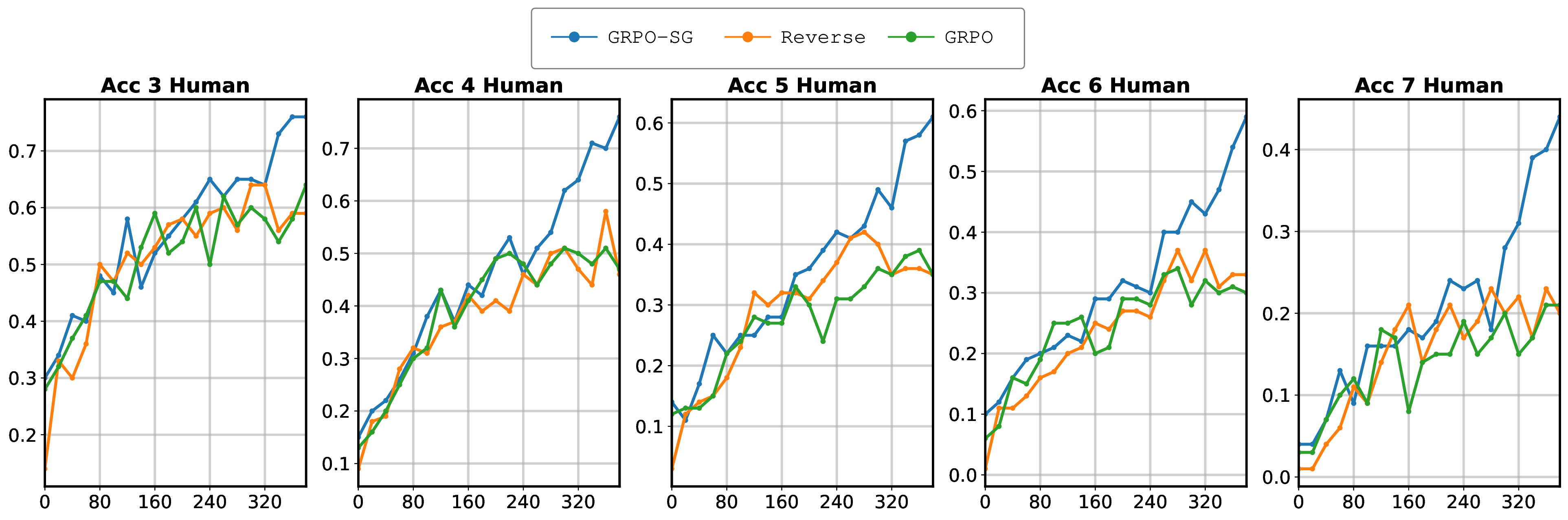}
    \caption{Accuracy on the \texttt{K\&K} Logic Puzzles benchmark, broken down by puzzle size (3--7 people). GRPO-SG consistently achieves higher accuracy than GRPO across all difficulty levels, while the Reverse variant yields performance comparable to GRPO without clear improvement.}
    \label{fig:compare_reverse}
\end{figure*}

In addition to the main experiments, we compare:
(i) \text{GRPO-SG}, which applies our confidence-aware token weighting rule;
(ii) a \text{Reverse} variant that flips this rule by assigning $2-w$ to any token whose GRPO-SG weight is $w$; and
(iii) the \text{GRPO} baseline. As shown in Figure~\ref{fig:compare_reverse}, \textcolor{blue}{\textbf{GRPO-SG (blue)}} consistently outperforms both
\textcolor{orange}{\textbf{Reverse (orange)}} and the \textcolor{green}{\textbf{GRPO baseline (green)}} across K\&K puzzle sizes (3--7),
while Reverse tracks GRPO closely without clear improvement. Beyond accuracy, this ablation supports our generalization view: GRPO-SG yields a more stable performance trajectory, while GRPO and the Reverse variant exhibit larger fluctuations.

\section{Proof of Our Theories}

\subsection{Proof of Lemma \ref{lem:optimal_solution}}

We have
\begin{align}
\pi_{*}= & \text{argmax}_{\pi}\left\{ \mathbb{E}_{q}\left[\sum_{t=1}^{|o|}\mathbb{E}_{o_{t}\sim\pi\left(\cdot\mid q,o_{<t}\right)}\left[r_{*}\left(\left[q,o_{<t}\right],o_{t}\right)\right]\right]\right.\nonumber \\
 & \left.-\lambda\sum_{t=1}^{|o|}D_{f}\left(\pi\left(\cdot\mid q,o_{<t}\right)\Vert\pi_{old}\left(\cdot\mid q,o_{<t}\right)\right)\right\} ,\label{eq:pi_r_appendix}
\end{align}

For $t\leq T$, we investigate
\begin{equation}
\max_{\pi}\mathbb{E}_{q}\left[\pi\left(o_{t}\mid q,o_{<t}\right)r_{*}\left([q,o_{<t}],o_{t}\right)\right]-\lambda D_{f}\left(\pi\left(\cdot\mid q,o_{<t}\right)\Vert\pi_{old}\left(\cdot\mid q,o_{<t}\right)\right).\label{eq:opt_t}
\end{equation}

We construct the Lagrange function
\begin{align*}
\mathcal{L}\left(\pi,\beta,\alpha\right) & =\mathbb{E}_{q}\left[\pi\left(o_{t}\mid q,o_{<t}\right)r_{*}\left([q,o_{<t}],o_{t}\right)\right]-\sum_{o_{t}}f\left(\frac{\pi\left(o_{t}\mid q,o_{<t}\right)}{\pi_{old}\left(o_{t}\mid q,o_{<t}\right)}\right)\pi_{old}\left(o_{t}\mid q,o_{<t}\right)\\
 & +\sum_{o_{t}}\alpha\left(o_{t}\right)\pi\left(o_{t}\mid q,o_{<t}\right)+\beta\left(\sum_{o_{t}}\pi\left(o_{t}\mid q,o_{<t}\right)-1\right).
\end{align*}

Using KKT conditions, we have
\begin{align*}
 & \frac{d\mathcal{L}}{d\pi\left(o_{t}\mid q,o_{<t}\right)}=r_{*}\left([q,o_{<t}],o_{t}\right)-\lambda f'\left(\frac{\pi\left(o_{t}\mid q,o_{<t}\right)}{\pi_{old}\left(o_{t}\mid q,o_{<t}\right)}\right)+\alpha\left(o_{t}\right)+\beta=0\\
 & \sum_{o_{t}}\pi\left(o_{t}\mid q,o_{<t}\right)-1=0\\
 & \sum_{o_{t}}\alpha\left(o_{t}\right)\geq0,\alpha\left(o_{t}\right)\pi\left(o_{t}\mid q,o_{<t}\right)=0
\end{align*}
We have $\pi\left(o_{t}\mid q,o_{<t}\right) >0$, leading to $\alpha(o_t)=0$ and
\[
r_{*}\left([q,o_{<t}],o_{t}\right)=\lambda f'\left(\frac{\pi\left(o_{t}\mid q,o_{<t}\right)}{\pi_{old}\left(o_{t}\mid q,o_{<t}\right)}\right)-\beta=\lambda f'\left(\frac{\pi\left(o_{t}\mid q,o_{<t}\right)}{\pi_{old}\left(o_{t}\mid q,o_{<t}\right)}\right)+const.
\]
By choosing 
\[
f\left(t\right)=\lambda^{-1}\int_{0}^{t}\frac{\omega\left(x\pi_{old}\left(o_{t}\right)\right)}{\pi_{old}\left(o_{t}\right)}dx,
\]
we have $f$ is convex and $f'\left(t\right)=\lambda^{-1}\frac{\omega\left(t\pi_{old}\left(o_{t}\right)\right)}{\pi_{old}\left(o_{t}\right)}$. 

Finally, consider $t=\frac{\pi_{*}\left(o_{t}\right)}{\pi_{old}\left(o_{t}\right)}$, we reach
\[
r_{*}\left([q,o_{<t}],o_{t}\right)=\lambda f'\left(\frac{\pi\left(o_{t}\right)}{\pi_{old}\left(o_{t}\right)}\right)+const=\frac{\omega\left(\pi_{*}\left(o_{t}\right)\right)}{\pi_{old}\left(o_{t}\right)}+const
\]

\subsection{Proof of Theorem  \ref{thm:shift}}\label{sec:shift}

We consider
\begin{align}
\max_{\theta} \; & \mathbb{E}_\mathcal{Q}\left[\frac{1}{|o|}\sum_{t=1}^{|o|}\mathbb{E}_{o_{\leq t}\sim\pi_{*}^{t}\left(\cdot\mid q\right)}\left[\omega\left(\pi_{*}\left(o_{t}\right)\right)\frac{\pi_{\theta}\left(o_{t}\right)}{\pi_{old}\left(o_{t}\right)}\right]\right] \nonumber\\
 & -\beta\mathbb{D}_{KL}\left(\pi_{\theta}\Vert\pi_{ref}\right).\label{eq:gen_new_appendix}
\end{align}

We rewrite the first term in the objective function as
\begin{align}
 & \mathbb{E}_{\mathcal{Q}}\left[\frac{1}{|o|}\sum_{t=1}^{|o|}\mathbb{E}_{o_{\leq t}\sim\pi_{*}^{t}\left(\cdot\mid q\right)}\left[\omega\left(\pi_{*}\left(o_{t}\right)\right)\frac{\pi_{\theta}\left(o_{t}\right)}{\pi_{old}\left(o_{t}\right)}\right]\right]\nonumber\\
= & \mathbb{E}_{\mathcal{Q}}\left[\frac{1}{|o|}\sum_{t=1}^{|o|}\sum_{o_{\leq t}}\pi_{*}^{t}\left(o_{\leq t}\mid q\right)\omega\left(\pi_{*}\left(o_{t}\right)\right)\frac{\pi_{\theta}\left(o_{t}\right)}{\pi_{old}\left(o_{t}\right)}\right]\nonumber \\
= & \mathbb{E}_{\mathcal{Q}}\left[\frac{1}{|o|}\sum_{t=1}^{|o|}\sum_{o_{\leq t}}\pi_{old}^{t}\left(o_{\leq t}\mid q\right)\omega\left(\pi_{\theta}\left(o_{t}\right)\right)\frac{\pi_{\theta}\left(o_{t}\right)}{\pi_{old}\left(o_{t}\right)}\right]\nonumber \\
+ & \mathbb{E}_{\mathcal{Q}}\left[\frac{1}{|o|}\sum_{t=1}^{|o|}\sum_{o_{\leq t}}\left[\pi_{*}^{t}\left(o_{\leq t}\mid q\right)-\pi_{old}^{t}\left(o_{\leq t}\mid q\right)\right]\omega\left(\pi_{*}\left(o_{t}\right)\right)\frac{\pi_{\theta}\left(o_{t}\right)}{\pi_{old}\left(o_{t}\right)}\right]\nonumber \\
+ & \mathbb{E}_{\mathcal{Q}}\left[\frac{1}{|o|}\sum_{t=1}^{|o|}\sum_{o_{\leq t}}\pi_{old}^{t}\left(o_{\leq t}\mid q\right)\left[\omega\left(\pi_{*}\left(o_{t}\right)\right)-\omega\left(\pi_{\theta}\left(o_{t}\right)\right)\right]\frac{\pi_{\theta}\left(o_{t}\right)}{\pi_{old}\left(o_{t}\right)}\right].\label{eq:sum} 
\end{align}
We further bound
\begin{align}
 & \mathbb{E}_{\mathcal{Q}}\left[\frac{1}{|o|}\sum_{t=1}^{|o|}\sum_{o_{\leq t}}\left(\pi_{*}^{t}\left(o_{\leq t}\mid q\right)-\pi_{old}^{t}\left(o_{\leq t}\mid q\right)\right)\omega\left(\pi_{*}\left(o_{t}\right)\right)\frac{\pi_{\theta}\left(o_{t}\right)}{\pi_{old}\left(o_{t}\right)}\right]\nonumber \\
\leq & \mathbb{E}_{\mathcal{Q}}\left[\frac{1}{|o|}\sum_{t=1}^{|o|}\left\{ \sum_{o_{\leq t}}\frac{\pi_{*}^{t}\left(o_{\leq t}\mid q\right)-\pi_{old}^{t}\left(o_{\leq t}\mid q\right)}{\pi_{*}^{t}\left(o_{\leq t}\mid q\right)}\pi_{*}^{t}\left(o_{\leq t}\mid q\right)^{1/2}\right\} \left\{ \pi_{*}^{t}\left(o_{\leq t}\mid q\right)^{1/2}\omega\left(\pi_{*}\left(o_{t}\right)\right)\frac{\pi_{\theta}\left(o_{t}\right)}{\pi_{old}\left(o_{t}\right)}\right\} \right]\nonumber \\
\stackrel{(1)}{\leq} & \mathbb{E}_{\mathcal{Q}}\left[\frac{1}{|o|}\sum_{t=1}^{|o|}\left\{ \sum_{o_{\leq t}}\pi_{*}^{t}\left(o_{\leq t}\mid q\right)\left(\frac{\pi_{*}^{t}\left(o_{\leq t}\mid q\right)-\pi_{old}^{t}\left(o_{\leq t}\mid q\right)}{\pi_{*}^{t}\left(o_{\leq t}\mid q\right)}\right)^{2}\right\} ^{\frac{1}{2}}\left\{ \sum_{o_{\leq t}}\pi_{*}^{t}\left(o_{\leq t}\mid q\right)\omega^{2}\left(\pi_{*}\left(o_{t}\right)\right)\left(\frac{\pi_{\theta}\left(o_{t}\right)}{\pi_{old}\left(o_{t}\right)}\right)^{2}\right\} ^{\frac{1}{2}}\right]\nonumber \\
\leq & A\mathbb{E}_{\mathcal{Q}}\left[\frac{1}{|o|}\sum_{t=1}^{|o|}D_{u}\left(\pi_{old}^{t}\left(o_{\leq t}\mid q\right)\|\pi_{*}^{t}\left(o_{\leq t}\mid q\right)\right)^{\frac{1}{2}}\left(\sum_{o_{\leq t}}\pi_{*}^{t}\left(o_{\leq t}\mid q\right)\left(\frac{\pi_{\theta}\left(o_{t}\right)}{\pi_{old}\left(o_{t}\right)}\right)^{2}\right)^{\frac{1}{2}}\right]\nonumber \\
= & \mathbb{E}_{\mathcal{Q}}\left[\frac{1}{|o|}\sum_{t=1}^{|o|}d\left(\pi_{old}^{t},\pi_{*}^{t}\right)\right],\label{eq:bound1}
\end{align}
where in $\stackrel{(1)}{\leq}$, we use Cauchy-Schwarz inequality, $D_u$ is a $f$-divergence with $u\left(t\right)=\left(t-1\right)^{2}$, $A$ is an upper-bound of $|\omega(\cdot)|$, and we define
\[
d\left(\pi_{old}^{t},\pi_{*}^{t}\right)=A\left(\sum_{o_{\leq t}}\pi_{*}^{t}\left(o_{\leq t}\mid q\right)\left(\frac{\pi_{\theta}\left(o_{t}\right)}{\pi_{old}\left(o_{t}\right)}\right)^{2}\right)^{\frac{1}{2}}D_{u}\left(\pi_{old}^{t}\left(o_{\leq t}\mid q\right)\|\pi_{*}^{t}\left(o_{\leq t}\mid q\right)\right)^{\frac{1}{2}}.
\]

\begin{align}
 & \mathbb{E}_{\mathcal{Q}}\left[\frac{1}{|o|}\sum_{t=1}^{|o|}\sum_{o_{\leq t}}\pi_{old}^{t}\left(o_{\leq t}\mid q\right)\left[\omega\left(\pi_{*}\left(o_{t}\right)\right)-\omega\left(\pi_{\theta}\left(o_{t}\right)\right)\right]\frac{\pi_{\theta}\left(o_{t}\right)}{\pi_{old}\left(o_{t}\right)}\right]\nonumber \\
= & \mathbb{E}_{\mathcal{Q}}\left[\frac{1}{|o|}\sum_{t=1}^{|o|}\left\{ \sum_{o_{\leq t}}\pi_{old}^{t}\left(o_{\leq t}\mid q\right)^{\frac{1}{2}}\left[\omega\left(\pi_{*}\left(o_{t}\right)\right)-\omega\left(\pi_{\theta}\left(o_{t}\right)\right)\right]\right\} \left\{ \pi_{old}^{t}\left(o_{\leq t}\mid q\right)^{\frac{1}{2}}\frac{\pi_{\theta}\left(o_{t}\right)}{\pi_{old}\left(o_{t}\right)}\right\} \right]\nonumber \\
\stackrel{(2)}{\leq} & \mathbb{E}_{\mathcal{Q}}\left[\frac{1}{|o|}\sum_{t=1}^{|o|}\left[\sum_{o_{\leq t}}\pi_{old}^{t}\left(o_{\leq t}\mid q\right)\left[\omega\left(\pi_{*}\left(o_{t}\right)\right)-\omega\left(\pi_{\theta}\left(o_{t}\right)\right)\right]^{2}\right]^{\frac{1}{2}}\left[\sum_{o_{\leq t}}\pi_{old}^{t}\left(o_{\leq t}\mid q\right)\left(\frac{\pi_{\theta}\left(o_{t}\right)}{\pi_{old}\left(o_{t}\right)}\right)^{2}\right]^{\frac{1}{2}}\right]\nonumber \\
= & \mathbb{E}_{\mathcal{Q}}\left[\frac{1}{|o|}\sum_{t=1}^{|o|}d'\left(\pi_{*}^{t},\pi_{\theta}^{t}\right)\right],\label{eq:bound2}
\end{align}
where in $\stackrel{(2)}{\leq}$, we use Cauchy-Schwarz inequality and we define
\[
d'\left(\pi_{*}^{t},\pi_{\theta}^{t}\right)=\left[\sum_{o_{\leq t}}\pi_{old}^{t}\left(o_{\leq t}\mid q\right)\left(\frac{\pi_{\theta}\left(o_{t}\right)}{\pi_{old}\left(o_{t}\right)}\right)^{2}\right]^{\frac{1}{2}}\left[\sum_{o_{\leq t}}\pi_{old}^{t}\left(o_{\leq t}\mid q\right)\left[\omega\left(\pi_{*}\left(o_{t}\right)\right)-\omega\left(\pi_{\theta}\left(o_{t}\right)\right)\right]^{2}\right]^{\frac{1}{2}}.
\]
Combining (\ref{eq:sum}), (\ref{eq:bound1}), and (\ref{eq:bound2}), we reach the conclusion.

\subsection{Proof of Theorem \ref{thm:gen_bound}}
We use the PAC-Bayes theory in this proof. In PAC-Bayes theory, $\theta$ could follow a distribution, says $P$, thus we define the expected loss over $\theta$ distributed by $P$ as follows:
\begin{align*}
    \Lc_{\Dc}(\theta,P) &= \mathbb{E}_{\theta\sim P}\big[\Lc_{\Dc}(\theta) \big] \\
    \Lc_{\Sc}(\theta,P) &= \E_{\theta\sim P}\big[\Lc_{\Sc}(\theta) \big].
\end{align*}
For any distribution $P= \mathcal{N}(\mathbf{0},\sigma_P^2\mathbb{I}_k)$ and $Q=\mathcal{N}(\theta,\sigma^2\mathbb{I}_k)$ over $\theta\in \mathbb{R}^k$, where $P$ is the prior distribution and $Q$ is the posterior distribution, use the PAC-Bayes theorem in \cite{PAC_Bayes}, for all $\beta>0$, with a probability at least $1-\delta$, we have
    \begin{equation}
        \Lc_{\Dc}(\theta,Q) \leq \Lc_{\Sc}(\theta,Q) +\frac{1}{\beta}\Big[\mathsf{KL}(Q\|P) + \log\frac{1}{\delta} + \Psi(\beta,N) \Big],\label{ineq:PAC-Bayes}
    \end{equation}
    where $\Psi$ is defined as
    \begin{align*}
        \Psi(\beta,N) = \log \E_{P}\E_{\Dc^N}\Big[\exp\big\{\beta\big[\Lc_{\Dc}(f_{\theta}) - \Lc_{\Sc}(f_{\theta})\big] \big\} \Big].
    \end{align*}
When the loss function is bounded by $L$, then 
    \begin{align*}
        \Psi(\beta,N) \leq \frac{\beta^2L^2}{8N}.
    \end{align*}
 The task is to minimize the second term of RHS of \eqref{ineq:PAC-Bayes},  we thus choose $\beta =\sqrt{8N} \frac{\mathsf{KL}(Q\|P) + \log\frac{1}{\delta}}{L}$. Then  the second term of RHS of \eqref{ineq:PAC-Bayes} is equal to
\begin{align*}
\sqrt{\frac{\mathsf{KL}(Q\|P) + \log \frac{1}{\delta}}{2N}}\times L.
\end{align*}
The KL divergence between $Q$ and $P$, when they are Gaussian, is given by formula
\begin{align*}
\mathsf{KL}(Q\|P) = \frac{1}{2}\left[ \frac{k\sigma^2 + \|\theta\|^2}{\sigma_P^2} - k + k \log\frac{\sigma_P^2}{\sigma^2}\right].
\end{align*}
For given posterior distribution $Q$ with fixed $\sigma^2$, to minimize the KL term, the $\sigma_P^2$ should be equal to $\sigma^2 + \|\theta\|^2/k$. In this case, the KL term is no less than 
\begin{align*}
 k \log\Big(1 +\frac{\|\theta\|^2}{k\sigma^2}  \Big).
\end{align*}
Thus, the second term of RHS is 
\begin{align*}
    \sqrt{\frac{\mathsf{KL}(Q\|P) + \log \frac{1}{\delta}}{2N}}\times L \geq \sqrt{\frac{k\log\big(1 + \frac{\|\theta\|^2}{k\sigma^2} \big)}{4N}}\times L \geq L
\end{align*}
when $\|\theta\|^2 > \sigma^2 \big\{\exp(4N/k)-1\big\}$. Hence, for any $\|\theta\|_2 > \sigma^2 \big\{\exp(4N/k)-1 \big\}$, we have the RHS is greater than the LHS, and the inequality is trivial. In this work, we only consider the case: 
\begin{align}\label{cond:theta}\|\theta\|^2 < \sigma^2 \big(\exp\{4N/k\} -1\big).
\end{align}
Distribution $P$ is Gaussian centered around  $\mathbf{0}$ with variance $\sigma_P^2 = \sigma^2 + \|\theta\|^2/k$, which is unknown at the time we set up the inequality, since $\theta$ is unknown. Meanwhile, we have to specify $P$ in advance, since $P$ is the prior distribution. To deal with this problem, we could choose a family of $P$ such that its means cover the space of $\theta$ satisfying inequality \eqref{cond:theta}. We set
\begin{align*}
c&= \sigma^2\big(1 + \exp\{4N/k\}\big)\\
P_j &= \mathcal{N}\big(0,c\exp\frac{1-j}{k}\mathbb{I}_k\big)\\
    \mathfrak{P}&:= \big\{P_j: j = 1,2,\ldots \big\}
\end{align*}
Then the following inequality holds for a particular distribution $P_j$ with probability $1-\delta_j$ with $\delta_j = \frac{6\delta}{\pi^2 j^2}$
\begin{align*}
    \mathbb{E}_{\theta^{\prime}\sim \mathcal{N}(\theta,\sigma^2)}\Lc_{\Dc}\big(f_{\theta^{\prime}} \big)&\leq \mathbb{E}_{\theta^{\prime}\sim \mathcal{N}(\theta,\sigma^2)} \Lc_{\Sc}\big(f_{\theta^{\prime}}\big) + \frac{1}{\beta}\left[  \mathsf{KL}(Q\|P_j) + \log\frac{1}{\delta_j} + \Psi(\beta,N) \right].
\end{align*}
Use the well-known equation: $\sum_{j=1}^{\infty} \frac{1}{j^2} = \frac{\pi^2}{6}$, then with probability $1-\delta$, the above inequality holds with every $j$. We pick
\begin{align*}
j^*:= \left\lfloor 1- k \log\frac{\sigma^2 + \|\theta\|^2/k}{c} \right\rfloor = \left\lfloor 1- k \log\frac{\sigma^2 + \|\theta\|^2/k}{\sigma^2(1+ \exp\{4N/k\})} \right\rfloor.
\end{align*}
Therefore,
\begin{align*}
&1 - j^* = \left\lceil k \log \frac{\sigma^2 + \|\theta\|^2/k}{c}\right\rceil \\
\Rightarrow \quad &\log \frac{\sigma^2 + \|\theta\|^2/k}{c}\leq \frac{1-j^*}{k} \leq \log\frac{\sigma^2 + \|\theta\|^2/k}{c} + \frac{1}{k}\\
\Rightarrow \quad & \sigma^2 + \|\theta\|^2/k \leq c\exp\left\{\frac{1-j^*}{k} \right\} \leq \exp(1/k) \big[\sigma^2 + \|\theta\|^2/k \big]\\
\Rightarrow \quad & \sigma^2 + \|\theta\|^2/k \leq \sigma_{P_{j^*}}^2\leq \exp(1/k) \big[\sigma^2 + \|\theta\|^2/k \big].
\end{align*}
Thus the KL term could be bounded as follow
\begin{align*}
    \mathsf{KL}(Q\|P_{j^*}) &= \frac{1}{2}\left[\frac{k\sigma^2 + \|\theta\|^2}{\sigma_{P_{j^*}}^2}- k + k \log \frac{\sigma_{P_{j^*}}^2}{\sigma^2} \right]\\
    &\leq \frac{1}{2}\left[\frac{k(\sigma^2 + \|\theta\|^2/k)}{\sigma^2 + \|\theta\|^2/k} - k + k \log \frac{\exp(1/k)\big(\sigma^2 + \|\theta\|^2/k\big)}{\sigma^2} \right] \\
    &= \frac{1}{2}\Big[k \log \frac{\exp(1/k)\big(\sigma^2 + \|\theta\|^2/k \big)}{\sigma^2} \Big] \\
    &= \frac{1}{2}\Big[1 + k\log\Big(1 + \frac{\|\theta\|^2}{k\sigma^2} \Big) \Big]
\end{align*}
For the term $\log \frac{1}{\delta_{j^*}}$, with recall that $c = \sigma^2\big(1+\exp(4N/k) \big)$ and $j^* = \left\lfloor 1- k \log\frac{\sigma^2 + \|\theta\|^2/k}{\sigma^2(1+ \exp\{4N/k\})} \right\rfloor$, we have
\begin{align*}
    \log\frac{1}{\delta_{j^*}} &= \log \frac{(j^*)^2\pi^2}{6\delta}  = \log\frac{1}{\delta}  + \log\Big(\frac{\pi^2}{6}\Big) + 2\log(j^*) \\
    &\leq \log\frac{1}{\delta} + \log\frac{\pi^2}{6} + 2\log \Big( 1+k\log\frac{\sigma^2\big(1+ \exp(4N/k)\big)}{\sigma^2 + \|\theta\|^2/k}\Big)  \\
    &\leq \log\frac{1}{\delta} + \log\frac{\pi^2}{6} + 2\log\Big(1+ k\log\big(1+\exp(4N/k)\big)\Big) \\
    &\leq \log\frac{1}{\delta} + \log\frac{\pi^2}{6} + 2\log\Big(1+ k\big(1+\frac{4N}{k} \big) \Big) \\
    &\leq \log\frac{1}{\delta} + \log\frac{\pi^2}{6} + \log(1+k + 4N).
\end{align*}
Hence, the inequality 
\begin{align*}
    \Lc_{\Dc}\Big(\theta^{\prime},\mathcal{N}(\theta,\sigma^2\mathbb{I}_k)\Big)&\leq \Lc_{\Sc}\Big(\theta^{\prime},\mathcal{N}(\theta,\sigma^2\mathbb{I}_k) \Big) + \sqrt{\frac{\mathsf{KL}(Q\|P_{j^*}) + \log \frac{1}{\delta_{j^*}}}{2N}}\times L \\
    &\leq \Lc_{\Sc}\Big(\theta^{\prime},\mathcal{N}(\theta,\sigma^2\mathbb{I}_k) \Big) \\ 
    & + \frac{L}{2\sqrt{N}}\sqrt{1 + k\log\Big(1+ \frac{\|\theta\|^2}{k\sigma^2}\Big) + 2 \log \frac{\pi^2}{6\delta} + 4 \log(N+k)}\\
    &\leq  \Lc_{\Sc}\Big(\theta^{\prime},\mathcal{N}(\theta,\sigma^2\mathbb{I}_k)\Big) \\ 
    & + \frac{L}{2\sqrt{N}} \sqrt{k\log\big(1+ \frac{\|\theta\|^2}{k\sigma^2}\big)+ O(1) + 2\log\frac{1}{\delta} + 4\log(N+k)}.
\end{align*}
Since $\|\theta^{\prime}-\theta\|^2$ is $k$ chi-square distribution, for any positive $t$,
we have
\begin{align*}
    \mathbb{P}\big(\|\theta^{\prime}-\theta\|^2 - k \sigma^2 \geq 2\sigma^2 \sqrt{kt} + 2t\sigma^2\big) \big) \leq \exp(-t).
\end{align*}
By choosing $t = \frac{1}{2}\log(N)$, with probability $1-N^{-1/2}$, we have
\begin{align*}
    \|\theta^{\prime}-\theta\|^2 \leq \sigma^2 \log(N) + k\sigma^2 + \sigma^2\sqrt{2 k\log(N)} \leq k\sigma^2 \Big(1 + \sqrt{\frac{\log(N)}{k}} \Big)^2.
\end{align*}
 By setting $\sigma = \rho\times \big(\sqrt{k} + \sqrt{\log(N)}\big)^{-1}$, we have $\|\theta^{\prime}-\theta\|^2 \leq \rho^2$. Hence, we get
 \begin{align*}
     \Lc_{\Sc}\Big(\theta^{\prime},\mathcal{N}(\theta,\sigma^2\mathbb{I}_k)\Big) &= \mathbb{E}_{\theta\sim\mathcal{N}(\theta,\sigma^2\mathbb{I}_k)}\mathbb{E}_{\Sc}\big[f_{\theta^{\prime}} \big] = \int_{\|\theta^{\prime}-\theta\|\leq \rho} \mathbb{E}_{\Sc}\big[f_{\theta^{\prime}}\big]d\mathcal{N}(\theta,\sigma^2\mathbb{I}) \\ 
     & + \int_{\|\theta^{\prime}-\theta\|> \rho} \mathbb{E}_{\Sc}\big[f_{\theta^{\prime}} \big] d\mathcal{N}(\theta,\sigma^2\mathbb{I})\\
     &\leq \Big(1-\frac{1}{\sqrt{N}} \Big)\max_{\|\theta^{\prime} - \theta\|\leq \rho} \Lc_{\Sc}(\theta^{\prime}) + \frac{1}{\sqrt{N}}L \\
     &\leq \max_{\|\theta^{\prime}-\theta\|_2 \leq \rho} \Lc_{\Sc}(\theta^{\prime}) + \frac{2L}{\sqrt{N}}.
 \end{align*}
 It follows that
\begin{align*}
    \Lc_{\Dc}(\theta) \leq \max_{\|\theta^{\prime}-\theta\|\leq \rho} \Lc_{\Sc}(\theta^{\prime}) & +\frac{4L}{\sqrt{N}}\Bigg[\sqrt{k\log\Big(1 + \frac{\|\theta\|^2}{\rho^2} \big(1+\sqrt{\log(N)/k}\big)^2 \Big)} \\ 
    & + 2\sqrt{\log\big(\frac{N+k}{\delta}\big)} + O(1) \Bigg]\\
    &= \Lc_\Dc(\theta \mid \Sc) + \frac{4L}{\sqrt{N}}\Bigg[\sqrt{k\log\Big(1 + \frac{\|\theta\|^2}{\rho^2} \big(1+\sqrt{\log(N)/k}\big)^2 \Big)}\\
    & + 2\sqrt{\log\big(\frac{N+k}{\delta}\big)} + O(1) \Bigg].
\end{align*}
For our case, the loss is bounded by $\left(1+\epsilon_{h}\right)\max\hat{A}_{i,t}\leq\left(1+\epsilon_{h}\right)G^{1/2}=L$. The reason is that $\frac{1}{G}\sum_{i=1}^{G}\hat{A}_{i,t}=0$ and $\frac{1}{G}\sum_{i=1}^{G}\hat{A}_{i,t}^{2}=1$. 

\subsection{Proof of Lemma \ref{lem:gradient}}

Let $\bar r_{i,t}(\theta)=w_{i,t}\frac{\pi_{\theta}(o_{i,t})}{\pi_{\mathrm{old}}(o_{i,t})}$. Because $w_{i,t}$ is computed with the stop-gradient operator in Eq.~(\ref{eq:token_weight}), it is treated as a constant when differentiating the surrogate. For $\hat{A}_{i,t} >0$, the inner term of the sum relevant to $o_{i,t}$ reduces to
\[
h\left(o_{i,t}\right)=\begin{cases}
\left(1+\epsilon_{h}\right)\hat{A}_{i,t} & \bar r_{i,t}(\theta)>1+\epsilon_{h}\\
\bar r_{i,t}(\theta)\hat{A}_{i,t} & \bar r_{i,t}(\theta)\leq1+\epsilon_{h}
\end{cases}
\]
The derivative w.r.t. $\theta$ becomes
\[
\nabla_{\theta}h\left(o_{i,t}\right)=\begin{cases}
0 & \bar r_{i,t}(\theta)>1+\epsilon_{h}\\
\bar r_{i,t}(\theta)\nabla_{\theta}\log\pi_{\theta}(o_{i,t})\hat{A}_{i,t} & \bar r_{i,t}(\theta)\leq1+\epsilon_{h}
\end{cases}
\]
Combining with the KL term derivative, we gain
\begin{align*}
 & \nabla_{\theta}h\left(o_{i,t}\right)+\beta w_{i,t}\pi_{ref}\left(o_{i,t}\right)\frac{\nabla_{\theta}\pi_{\theta}\left(o_{i,t}\right)}{\pi_{\theta}\left(o_{i,t}\right)^{2}}-\beta w_{i,t}\nabla_{\theta}\log\pi_{\theta}(o_{i,t})\\
= & \nabla_{\theta}h\left(o_{i,t}\right)+\beta w_{i,t}\frac{\pi_{ref}\left(o_{i,t}\right)}{\pi_{\theta}\left(o_{i,t}\right)}\nabla_{\theta}\log\pi_{\theta}\left(o_{i,t}\right)-\beta w_{i,t}\nabla_{\theta}\log\pi_{\theta}(o_{i,t})
\end{align*}

For $\hat{A}_{i,t} <0$, the inner term of the sum relevant to $o_{i,t}$ reduces to
\[
h\left(o_{i,t}\right)=\begin{cases}
\left(1-\epsilon_{l}\right)\hat{A}_{i,t} & \bar r_{i,t}(\theta)<1-\epsilon_{l}\\
\bar r_{i,t}(\theta)\hat{A}_{i,t} & \bar r_{i,t}(\theta)\geq1-\epsilon_{l}
\end{cases}
\]
The derivative w.r.t. $\theta$ becomes
\[
\nabla_{\theta}h\left(o_{i,t}\right)=\begin{cases}
0 & \bar r_{i,t}(\theta)<1-\epsilon_{l}\\
\bar r_{i,t}(\theta)\nabla_{\theta}\log\pi_{\theta}(o_{i,t})\hat{A}_{i,t} & \bar r_{i,t}(\theta)\geq1-\epsilon_{l}
\end{cases}
\]
Combining with the KL term derivative, we gain
\begin{align*}
 & \nabla_{\theta}h\left(o_{i,t}\right)+\beta w_{i,t}\pi_{ref}\left(o_{i,t}\right)\frac{\nabla_{\theta}\pi_{\theta}\left(o_{i,t}\right)}{\pi_{\theta}\left(o_{i,t}\right)^{2}}-\beta w_{i,t}\nabla_{\theta}\log\pi_{\theta}(o_{i,t})\\
= & \nabla_{\theta}h\left(o_{i,t}\right)+\beta w_{i,t}\frac{\pi_{ref}\left(o_{i,t}\right)}{\pi_{\theta}\left(o_{i,t}\right)}\nabla_{\theta}\log\pi_{\theta}\left(o_{i,t}\right)-\beta w_{i,t}\nabla_{\theta}\log\pi_{\theta}(o_{i,t})
\end{align*}

Finally, leveraging the two above cases, we gain the final formula. The same clipping decision can be written compactly as $\mathbb{I}_{\mathrm{sharp}}=\mathbf{1}\!\left[\bar r_{i,t}(\theta)\hat{A}_{i,t}\leq\operatorname{clip}\!\left(\bar r_{i,t}(\theta),1-\epsilon_l,1+\epsilon_h\right)\hat{A}_{i,t}\right]$, or equivalently on the original likelihood ratio with thresholds rescaled by $w_{i,t}^{-1}$.

\begin{lemma}
    \label{lem:min_max}
    Assume that $A_{1:m}$ is a sequence of matrices with $\sigma_{\text{min}}(A_i) \geq a_i^2 \geq 0$ and $\sigma_{\text{max}}(A_i) \leq b_i^2$ for all $i \in \{1,...,m\}$. We then have 
    \[
\Vert x\Vert_{2}\prod_{i=m}^{1}a_{i}\leq\Vert x\prod_{i=m}^{1}A_{i}\Vert_{2}\leq\Vert x\Vert_{2}\prod_{i=m}^{1}b_{i}
\]
\begin{proof}
We start with
\[
\Vert xA\Vert_{2}^{2}=xAA^{T}x^{T}.
\]

According to the Rayleigh inequality, we have
\[
\sigma_{\text{min}}\left(A\right)\Vert x\Vert_{2}^{2}\leq\Vert xA\Vert_{2}^{2}=xAA^{T}x^{T}\leq\sigma_{\text{max}}\left(A\right)\Vert x\Vert_{2}^{2}
\]
\begin{equation}
\sigma_{\text{min}}\left(A\right)^{1/2}\Vert x\Vert_{2}\leq\Vert xA\Vert_{2}\leq\sigma_{\text{max}}\left(A\right)^{1/2}\Vert x\Vert_{2}\label{eq:norm_xA}
\end{equation}

Consider $A=\prod_{i=m}^{1}A_{i}$ and apply Inequality \ref{eq:norm_xA} recursively, we obtain
\[
\Vert x\Vert_{2}\prod_{i=m}^{1}\sigma_{\text{min}}\left(A_{i}\right)^{1/2}\leq\Vert x\prod_{i=m}^{1}A_{i}\Vert_{2}\leq\Vert x\Vert_{2}\prod_{i=m}^{1}\sigma_{\text{max}}\left(A\right)^{1/2}
\]
\[
\Vert x\Vert_{2}\prod_{i=m}^{1}a_{i}\leq\Vert x\prod_{i=m}^{1}A_{i}\Vert_{2}\leq\Vert x\Vert_{2}\prod_{i=m}^{1}b_{i}
\]
    
\end{proof}

\end{lemma}

\subsection{Proof of Theorem \ref{thm:bounds}}

We first have
\[
\frac{\partial\log\pi_{\theta}\left(o_{i,t}\right)}{\partial h_{i,t}}=1_{k}-p_{i,t},
\]
where $1_k$ is a one-hot vector over the vocabulary, the token $o_{i,t}$ has the index $k$ in the vocabulary, $p_{i,t}$ is the distribution over vocabulary with $\pi_{\theta}\left(o_{i,t}\right)= p_{i,t}(k)$.

Consider a specific layer $l$, we then have
\begin{align*}
\frac{\partial\log\pi_{\theta}\left(o_{i,t}\right)}{\partial\theta_{l}} & =\frac{\partial\log\pi_{\theta}\left(o_{i,t}\right)}{\partial h_{i,t}}\frac{\partial h_{i,t}}{\partial a_{L}}\frac{\partial a_{L}}{\partial a_{L-1}}...\frac{\partial a_{l+1}}{\partial a_{l}}\frac{\partial a_{l}}{\partial\theta_{l}}\\
= & \left[1_{k}-p_{i,t}\right]W\prod_{i=l}^{L}J_{i}G_{l}.
\end{align*}

Applying Lemma \ref{lem:min_max}, we reach
\[
\Vert1_{k}-p_{i,t}\Vert_{2}a^{W}a_{l}^{G}\prod_{i=l}^{L}a_{i}^{J}\leq\Vert\frac{\partial\log\pi_{\theta}\left(o_{i,t}\right)}{\partial\theta_{l}}\Vert_{2}\leq\Vert1_{k}-p_{i,t}\Vert_{2}b^{W}b_{l}^{G}\prod_{i=l}^{L}b_{i}^{J}
\]

We further bound
\[
\Vert1_{k}-p_{i,t}\Vert_{2}=\sqrt{\left(1-p_{i,t}\left(k\right)\right)^{2}+\sum_{j\neq k}p_{i,t}\left(j\right)^{2}}\geq1-p_{i,t}\left(k\right)=1-\pi_{\theta}\left(o_{i,t}\right).
\]
\begin{align*}
\Vert1_{k}-p_{i,t}\Vert_{2} & =\sqrt{\left(1-p_{i,t}\left(k\right)\right)^{2}+\sum_{j\neq k}p_{i,t}\left(j\right)^{2}}\leq\sqrt{\left(1-p_{i,t}\left(k\right)\right)^{2}+\left(\sum_{j\neq k}p_{i,t}\left(j\right)\right)^{2}}\\
= & \sqrt{\left(1-p_{i,t}\left(k\right)\right)^{2}+\left(1-p_{i,t}\left(k\right)\right)^{2}}=\sqrt{2}\left(1-p_{i,t}\left(k\right)\right)=\sqrt{2}\left(1-\pi_{\theta}\left(o_{i,t}\right)\right).
\end{align*}
Therefore, we further reach
\[
\left(1-\pi_{\theta}\left(o_{i,t}\right)\right)a^{W}a_{l}^{G}\prod_{i=l}^{L}a_{i}^{J}\leq\Vert\frac{\partial\log\pi_{\theta}\left(o_{i,t}\right)}{\partial\theta_{l}}\Vert_{2}\leq\sqrt{2}\left(1-\pi_{\theta}\left(o_{i,t}\right)\right)b^{W}b_{l}^{G}\prod_{i=l}^{L}b_{i}^{J}
\]
Finally, by noting that
\[
\Vert\nabla_{\theta}\log\pi_{\theta}\left(o_{i,t}\right)\Vert_{2}^{2}=\sum_{l=1}^{L}\Vert\frac{\partial\log\pi_{\theta}\left(o_{i,t}\right)}{\partial\theta_{l}}\Vert_{2}^{2},
\] hence leading to
\[
\frac{1}{\sqrt{L}}\sum_{l=1}^{L}\Vert\frac{\partial\log\pi_{\theta}\left(o_{i,t}\right)}{\partial\theta_{l}}\Vert_{2}\leq\Vert\nabla_{\theta}\log\pi_{\theta}\left(o_{i,t}\right)\Vert_{2}\leq\sum_{l=1}^{L}\Vert\frac{\partial\log\pi_{\theta}\left(o_{i,t}\right)}{\partial\theta_{l}}\Vert_{2}
\]

Finally, we have 
\[
\Vert g_{i,t}\Vert_{2}=w_{it}\left|\gamma_{i,t}\right|\Vert\nabla_{\theta}\log\pi_{\theta}(o_{i,t})\Vert_{2}.
\]
Using the bound for $\Vert\nabla_{\theta}\log\pi_{\theta}(o_{i,t})\Vert_{2}$ developed above, we reach the conclusion.

To prove the theorem \ref{thm:grad_bound}, we need the following lemma.
\begin{lemma} \label{lem:almost orthogonal}
Let $v_{1},...,v_{n}\stackrel{iid}{\sim}\text{Unif}\left(\mathbb{S}^{d-1}\right)\subset\mathbb{R}^{d}$. There exists universal constants $C,c >0$ such that for $\delta \in (0,1)$, with probability at least $1-\delta$, we have
\begin{equation}
\max_{i\neq j}\left|\left\langle v_{i},v_{j}\right\rangle \right|
\leq
\sqrt{\frac{1}{cd}\left(\log\left(Cn^{2}\right)+\log\frac{1}{\delta}\right)}.
\label{eq:max_bound}
\end{equation}
\end{lemma}
\begin{proof}
We consider the function $f_{u}\left(v\right)=\left\langle v,u\right\rangle$ with $v\in\mathbb{S}^{d-1}$. We are going to apply Levy's lemma to the function $f_u$. 

First, we verify the Lipschitz constraint. For any $v,v'\in\mathbb{S}^{d-1}$,  we have
\[
\left|f_{u}\left(v\right)-f_{u}\left(v'\right)\right|=\left|\left\langle v-v',u\right\rangle \right|\leq\|v-v'\|_{2}\|u\|_{2}=\|v-v'\|_{2},
\]
hence $f_{u}\left(v\right)$ is 1-Lipschitz on $\mathbb{S}^{d-1}$.

It is obvious that $\mathbb{E}_{v}\left[\left\langle v,u\right\rangle \right]=0$ and also $\text{med}(f_u(v)) = 0, v \sim \mathbb{S}^{d-1}$. 

Using Levy's lemma, for every $\epsilon>0$, we have
\[
\mathbb{P}\left(\left|\left\langle v,u\right\rangle \right|\geq\epsilon\right)\leq C\exp\left(-cd\epsilon^{2}\right).
\]

From the two-vector result, we have
\[
\mathbb{P}\left(\left|\left\langle v_{i},v_{j}\right\rangle \right|\geq\epsilon\right)\leq C\exp\left(-cd\epsilon^{2}\right),\forall i\neq j.
\]

This follows that
\[
\mathbb{P}\left(\exists i\neq j:\left|\left\langle v_{i},v_{j}\right\rangle \right|\geq\epsilon\right)\leq n^{2}C\exp\left(-cd\epsilon^{2}\right).
\]

To ensure $n^{2}C\exp\left(-cd\epsilon^{2}\right)\leq\delta$, we choose $\epsilon = \sqrt{\frac{1}{cd}\left(\log\left(Cn^{2}\right)+\log\frac{1}{\delta}\right)}$, leading to
\[
\mathbb{P}\left(\max_{i\neq j}\left|\left\langle v_{i},v_{j}\right\rangle \right|<\epsilon\right)\geq1-\delta.
\]

\end{proof}

\subsection{Proof of Theorem \ref{thm:grad_bound}}

(i) We have

\[
\nabla_{\theta}\mathcal{L}_{\mathcal{S}}(\pi_{\theta})=\mathbb{E}_{\mathcal{S}}\left[\frac{1}{\sum_{i}^{G}|o_{i}|}\sum_{i=1}^{G}\sum_{t=1}^{|o_{i}|}g_{i,t}\right].
\]

Therefore, we obtain

\[
\|\nabla_{\theta}\mathcal{L}_{\mathcal{S}}(\pi_{\theta})\|_{2}\leq\mathbb{E}_{\mathcal{S}}\left[\frac{1}{\sum_{i}^{G}|o_{i}|}\sum_{i=1}^{G}\sum_{t=1}^{|o_{i}|}\|g_{i,t}\|_{2}\right]\leq\mathbb{E}_{\mathcal{S}}\left[\frac{\sqrt{2}}{\sum_{i}^{G}|o_{i}|}\sum_{i=1}^{G}\sum_{t=1}^{|o_{i}|}w_{i,t}\left(1-\pi\left(o_{i,t}\right)\right)U_{i,t}\right].
\]

ii) We write the gradient $\nabla_{\theta}\mathcal{L}_{\mathcal{S}}(\pi_{\theta})$ as
\[
\nabla_{\theta}\mathcal{L}_{\mathcal{S}}(\pi_{\theta})=\mathbb{E}_{\mathcal{S}}\left[\frac{1}{\sum_{i=1}^{G}|o_{i}|}\sum_{i=1}^{G}\sum_{t=1}^{|o_{i}|}g_{i,t}\right]=\frac{1}{|\mathcal{S}|}\sum_{k=1}^{N}\frac{1}{\sum_{i=1}^{G}|o_{ki}|}\sum_{i=1}^{G}\sum_{t=1}^{|o_{ki}|}g_{i,t}^{k}.
\]
This follows that
\begin{align*}
 & \|\nabla_{\theta}\mathcal{L}_{\mathcal{S}}(\pi_{\theta})\|_{2}^{2}=\frac{1}{|\mathcal{S}|^{2}}\Vert\sum_{k=1}^{N}\frac{1}{\sum_{i}^{G}|o_{ki}|}\sum_{i=1}^{G}\sum_{t=1}^{|o_{ki}|}g_{i,t}^{k}\|_{2}^{2}\\
= & \frac{1}{|\mathcal{S}|^{2}}\left[\sum_{k=1}^{N}\sum_{i=1}^{G}\sum_{t=1}^{|o_{ki}|}\frac{\|g_{i,t}^{k}\|_{2}^{2}}{(\sum_{i=1}^{G}|o_{ki}|)^{2}}\right]+\frac{1}{|\mathcal{S}|^{2}}\sum_{k,k'}\sum_{i,i'}\sum_{t,t'}\frac{1}{(\sum_{i}|o_{ki}|)(\sum_{i'}|o_{k'i'}|)}\left\langle \frac{g_{i.t}^{k}}{\|g_{i,t}^{k}\|},\frac{g_{i',t'}^{k'}}{\|g_{i',t'}^{k'}\|}\right\rangle \|g_{i,t}^{k}\|\|g_{i',t'}^{k'}\|
\end{align*}
Using Lemma \ref{lem:almost orthogonal} for $v_{i,t}^{k}=\frac{g_{i.t}^{k}}{\|g_{i,t}^{k}\|}$, we have with probability at least $1-\delta$, we have
\[
\left\langle \frac{g_{i.t}^{k}}{\|g_{i,t}^{k}\|},\frac{g_{i',t'}^{k'}}{\|g_{i',t'}^{k'}\|}\right\rangle \leq\epsilon,
\]
with $\epsilon=\sqrt{\frac{1}{cd}\left(\log\left(CK^{2}\right)+\log\frac{1}{\delta}\right)}$ where $K=\sum_{k=1}^{N}\sum_{i=1}^{G}|o_{ki}|$. 

This further implies
\begin{align*}
 & \|\nabla_{\theta}\mathcal{L}_{\mathcal{S}}(\pi_{\theta})\|_{2}^{2}\geq\frac{1}{|\mathcal{S}|^{2}}\left[\sum_{k=1}^{N}\left(\frac{1}{(\sum_{i=1}^{G}|o_{ki}|)^{3/2}}\sum_{i=1}^{G}\sum_{t=1}^{|o_{ki}|}\|g_{i,t}^{k}\|_{2}\right)^{2}\right]-\frac{\epsilon}{|\mathcal{S}|^{2}}\sum_{k,k'}\sum_{i,i'}\sum_{t,t'}\frac{1}{(\sum_{i}|o_{ki}|)(\sum_{i'}|o_{k'i'}|)}\|g_{i,t}\|\|g_{i',t'}\|\\
\geq & \frac{1}{|\mathcal{S}|^{2}}\left[\frac{1}{|\mathcal{S}|}\left(\sum_{k=1}^{N}\frac{1}{(\sum_{i=1}^{G}|o_{ki}|)^{3/2}}\sum_{i=1}^{G}\sum_{t=1}^{|o_{ki}|}\|g_{i,t}^{k}\|_{2}\right)^{2}\right]-\frac{\epsilon}{|\mathcal{S}|^{2}}\left[\sum_{k}\sum_{i}\sum_{t}\frac{\|g_{i,t}\|}{\sum_{i}|o_{ki}|}\right]\left[\sum_{k'}\sum_{i'}\sum_{t'}\frac{\|g_{i',t'}\|}{\sum_{i}|o_{k'i'}|}\right]\\
\geq & \frac{1}{|\mathcal{S}|^{2}}\mathbb{E}_{\mathcal{S}}\left[\left(\frac{1}{(\sum_{i=1}^{G}|o_{i}|)^{3/2}}\sum_{i=1}^{G}\sum_{t=1}^{|o_{i}|}\|g_{i,t}\|_{2}\right)^{2}\right]-\epsilon\mathbb{E}_{\mathcal{S}\times\mathcal{S}}\left[\left(\sum_{i}\sum_{t}\frac{\|g_{i,t}\|}{\sum_{i}|o_{i}|}\right)\left(\sum_{i}\sum_{t}\frac{\|g_{i,t}^{'}\|}{\sum_{i}|o_{i}^{'}|}\right)\right]
\end{align*}

Finally, using the upper and lower bounds of $\|g_{i.t}\|_2$, we gain
\begin{align*}
\|\nabla_{\theta}\mathcal{L}_{\mathcal{S}}(\pi_{\theta})\|_{2}^{2} & \geq\frac{1}{|\mathcal{S}|^{2}}\mathbb{E}_{\mathcal{S}}\left[\left(\sum_{i=1}^{G}\sum_{t=1}^{|o_{i}|}\frac{w_{i,t}\left(1-\pi_{\theta}\left(o_{i,t}\right)\right)L_{i,t}}{(\sum_{i=1}^{G}|o_{i}|)^{3/2}\sqrt{L}}\right)^{2}\right]\\
 & -2\epsilon\mathbb{E}_{\mathcal{S}\times\mathcal{S}}\left[\left(\sum_{i=1}^{G}\sum_{t=1}^{|o_{i}|}\frac{w_{i,t}\left(1-\pi_{\theta}\left(o_{i,t}\right)\right)U_{i,t}}{\sum_{i}|o_{i}|}\right)\left(\sum_{i=1}^{G}\sum_{t=1'}^{|o_{i}^{'}|}\frac{w_{i,t}^{'}\left(1-\pi_{\theta}\left(o_{i,t}^{'}\right)\right)U_{i,t}^{'}}{\sum_{i'}|o_{i}^{'}|}\right)\right].
\end{align*}


\newpage
\section*{NeurIPS Paper Checklist}

\begin{enumerate}

\item {\bf Claims}
    \item[] Question: Do the main claims made in the abstract and introduction accurately reflect the paper's contributions and scope?
    \item[] Answer: \answerYes{}
    \item[] Justification: The abstract and introduction summarize the main methodological, theoretical, and empirical claims, and these claims are supported by Sections~1,~3, and~4.

\item {\bf Limitations}
    \item[] Question: Does the paper discuss the limitations of the work performed by the authors?
    \item[] Answer: \answerYes{}
    \item[] Justification: The Appendix section~\ref{subsec:limitations} discusses computational overhead as a limitation, and Appendix Section~\ref{subsec:comp-costs} quantifies the added cost of GRPO-SG.

\item {\bf Theory assumptions and proofs}
    \item[] Question: For each theoretical result, does the paper provide the full set of assumptions and a complete (and correct) proof?
    \item[] Answer: \answerYes{}
    \item[] Justification: The paper states the assumptions used by the main theoretical results in Section~3 and provides the corresponding derivations and proofs in the appendix.

    \item {\bf Experimental result reproducibility}
    \item[] Question: Does the paper fully disclose all the information needed to reproduce the main experimental results of the paper to the extent that it affects the main claims and/or conclusions of the paper (regardless of whether the code and data are provided or not)?
    \item[] Answer: \answerYes{}
    \item[] Justification: The appendix specifies datasets, prompts, reward functions, model backbones, hyperparameters, and evaluation protocols for the three RLVR settings in Section~\ref{app:exp-detail}.

\item {\bf Open access to data and code}
    \item[] Question: Does the paper provide open access to the data and code, with sufficient instructions to faithfully reproduce the main experimental results, as described in supplemental material?
    \item[] Answer: \answerNo{}
    \item[] Justification: The draft describes the setup and references a code release, but it does not yet include an anonymized open code-and-data package with exact reproduction commands in the supplemental material.

\item {\bf Experimental setting/details}
    \item[] Question: Does the paper specify all the training and test details (e.g., data splits, hyperparameters, how they were chosen, type of optimizer) necessary to understand the results?
    \item[] Answer: \answerYes{}
    \item[] Justification: Training details, evaluation settings, prompts, rewards, and the main hyperparameters are reported in Section~4 and Appendix Section~\ref{app:exp-detail}.

\item {\bf Experiment statistical significance}
    \item[] Question: Does the paper report error bars suitably and correctly defined or other appropriate information about the statistical significance of the experiments?
    \item[] Answer: \answerNo{}
    \item[] Justification: The current draft reports single performance numbers without confidence intervals, error bars, or formal significance tests.

\item {\bf Experiments compute resources}
    \item[] Question: For each experiment, does the paper provide sufficient information on the computer resources (type of compute workers, memory, time of execution) needed to reproduce the experiments?
    \item[] Answer: \answerYes{}
    \item[] Justification: The appendix reports the GPU type together with representative runtime and memory measurements for the training setup in Appendix Section~\ref{subsec:comp-costs}.
    
\item {\bf Code of ethics}
    \item[] Question: Does the research conducted in the paper conform, in every respect, with the NeurIPS Code of Ethics \url{https://neurips.cc/public/EthicsGuidelines}?
    \item[] Answer: \answerYes{}
    \item[] Justification: The work uses public benchmarks and standard LLM post-training procedures and does not involve human-subject data collection or deceptive experimental practices.

\item {\bf Broader impacts}
    \item[] Question: Does the paper discuss both potential positive societal impacts and negative societal impacts of the work performed?
    \item[] Answer: \answerNA{}
    \item[] Justification: There is no societal impact of the work performed.
    
\item {\bf Safeguards}
    \item[] Question: Does the paper describe safeguards that have been put in place for responsible release of data or models that have a high risk for misuse (e.g., pre-trained language models, image generators, or scraped datasets)?
    \item[] Answer: \answerNA{}
    \item[] Justification: The paper does not release model weights or a scraped high-risk dataset; it studies a training recipe on top of existing models and benchmarks.

\item {\bf Licenses for existing assets}
    \item[] Question: Are the creators or original owners of assets (e.g., code, data, models), used in the paper, properly credited and are the license and terms of use explicitly mentioned and properly respected?
    \item[] Answer: \answerNo{}
    \item[] Justification: The paper cites the datasets, benchmarks, and base models it uses, but it does not yet enumerate license names and terms of use for each external asset.

\item {\bf New assets}
    \item[] Question: Are new assets introduced in the paper well documented and is the documentation provided alongside the assets?
    \item[] Answer: \answerNA{}
    \item[] Justification: The current submission does not introduce a new public dataset or released model artifact as a standalone asset.

\item {\bf Crowdsourcing and research with human subjects}
    \item[] Question: For crowdsourcing experiments and research with human subjects, does the paper include the full text of instructions given to participants and screenshots, if applicable, as well as details about compensation (if any)? 
    \item[] Answer: \answerNA{}
    \item[] Justification: The paper does not involve crowdsourcing experiments or research with human subjects.

\item {\bf Institutional review board (IRB) approvals or equivalent for research with human subjects}
    \item[] Question: Does the paper describe potential risks incurred by study participants, whether such risks were disclosed to the subjects, and whether Institutional Review Board (IRB) approvals (or an equivalent approval/review based on the requirements of your country or institution) were obtained?
    \item[] Answer: \answerNA{}
    \item[] Justification: The paper does not involve crowdsourcing experiments or research with human subjects.

\item {\bf Declaration of LLM usage}
    \item[] Question: Does the paper describe the usage of LLMs if it is an important, original, or non-standard component of the core methods in this research? Note that if the LLM is used only for writing, editing, or formatting purposes and does \emph{not} impact the core methodology, scientific rigor, or originality of the research, declaration is not required.
    \item[] Answer: \answerNA{}
    \item[] Justification: LLMs were used only to help with writing and editing the manuscript, not as a non-standard component of the core scientific method.

\end{enumerate}

\end{document}